\documentclass[11pt]{article}

\usepackage[final]{acl}

\usepackage{latexsym}

\usepackage{iftex}
\ifXeTeX
  \usepackage{fontspec}
  \usepackage{ucharclasses}
  
  \newfontfamily\sinhalafont{NotoSansSinhala}[
  Path = ./font-NotoSansSinhala/,
  UprightFont = NotoSansSinhala-Regular.ttf,
  BoldFont = NotoSansSinhala-Bold.ttf,
  Renderer = HarfBuzz,
  Script = Sinhala,
  Scale = MatchLowercase
]

  \setTransitionTo{Sinhala}{\sinhalafont}
  \setTransitionFrom{Sinhala}{\rmfamily}
\else
  \usepackage{times}
  \usepackage[T1]{fontenc}
  \usepackage[utf8]{inputenc}
  \usepackage{inconsolata}
\fi

\usepackage{microtype}
\usepackage{graphicx}
\usepackage{xcolor}
\definecolor{editcol}{HTML}{D55E00}

\usepackage{amsmath}
\usepackage{amssymb}
\usepackage{booktabs}
\usepackage{multirow}
\usepackage{makecell}
\usepackage{tabularx}
\usepackage{colortbl}
\usepackage{subcaption}
\usepackage{tikz}
\usepackage{needspace}

\newcommand{\hf}[2]{\raisebox{-2.2pt}{\includegraphics[scale=0.09]{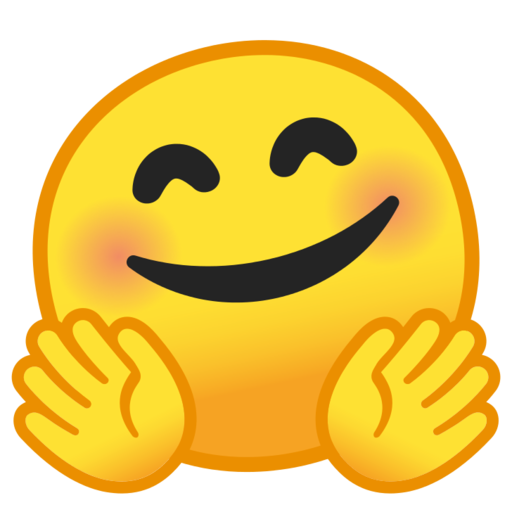}}~\href{#1}{\texttt{#2}}}

\newcommand{\gh}[2]{\raisebox{-2.2pt}{\includegraphics[scale=0.02]{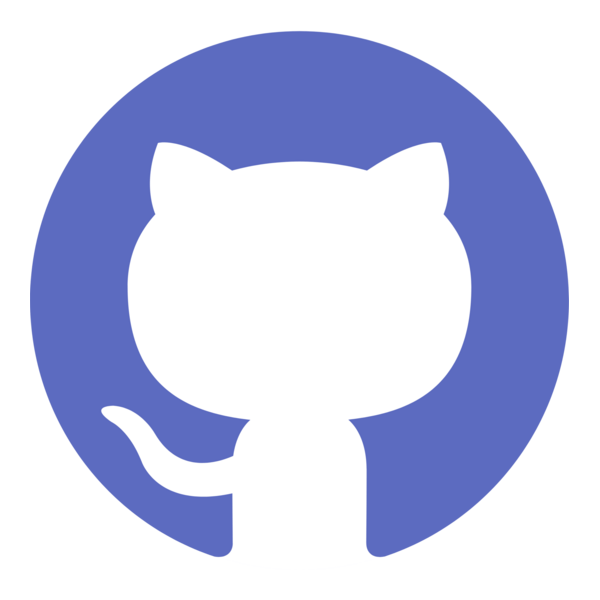}}~\href{#1}{\texttt{#2}}}

\usetikzlibrary{positioning,arrows.meta}

\newif\ifdraftnums
\draftnumsfalse
\newcommand{\true}[1]{\ifdraftnums{\color{blue}#1}\else#1\fi}

\title{Confident but Wrong: A Constrained Decoding Diagnostic\\for Low-Resource Automatic Post-Editing}

\author{
 \textbf{Isuru Wijesiri\textsuperscript{1}},
 \textbf{Nisansa de Silva\textsuperscript{2}},
 \textbf{Kavindu Warnakulasuriya\textsuperscript{3}}\\
 \textbf{Aloka Fernando\textsuperscript{2}},
  \textbf{Surangika Ranathunga\textsuperscript{4}}
\\
 \textsuperscript{1}WSO2, Sri Lanka \quad
 \textsuperscript{2}Dept.\ of CSE, University of Moratuwa, Sri Lanka \\
 \textsuperscript{3}National University of Singapore \quad
 \textsuperscript{4}Massey University, New Zealand
\\
 {\small \texttt{NisansaDdS@cse.mrt.ac.lk},
 \texttt{alokaf@cse.mrt.ac.lk},}\\
 {\small \texttt{kavindu\_warnakulasuriya@u.nus.edu},
 \texttt{s.ranathunga@massey.ac.nz}}\\
 {\small \textbf{Correspondence:} \texttt{imwijesiri@gmail.com}}
}

\begin{document}
\maketitle

\begin{abstract}

  Automatic Post-Editing (APE) for low-resource languages (LRLs) often fails
to improve Machine Translation (MT), and the score alone cannot say
why: whether more training would help, or whether the training data
is too inconsistent to learn from. We introduce a
  black-box, inference-time diagnostic that tells these two cases
  apart without retraining or annotation. It varies an edit-distance
  penalty $\lambda$ that drives the model from free editing towards
  copying the MT, and reads two signals: (1) the shape of the
  Translation Edit Rate (TER)-vs-$\lambda$ curve, U-shaped if edits from the model reduce error and monotonically decreasing if none does;
  and (2) the ordering of constraint variants that trust model confidence to increasing degrees, which shows whether confidence
  tracks edit quality. Across decoder-only and encoder--decoder models
  on English--Sinhala, the diagnostic exposes two failure modes
  consistent with a heterogeneous post-edit signal as the underlying
  cause: \textbf{Binary Collapse}, where the model copies the MT or
  makes off-target edits, and \textbf{Confident Miscalibration}, where
  the confidence signals we test do not separate useful edits from
  unnecessary ones. The pattern holds on English--Marathi and
  English--Tamil, with the failure modes tracking the post-edit
  distribution rather than MT quality or language family. Beyond
  diagnosis, the curve shape prescribes a concrete next step for
  practitioners; in the favorable case, a static constraint yields a
  free inference-time accuracy gain. We release the first
  English--Sinhala (${\sim}$66k) and a new English--Tamil (${\sim}$39k) APE
  datasets with all code.
\end{abstract}

\section{Introduction}
\label{sec:intro}

Automatic Post-Editing (APE) corrects residual errors in Machine
Translation (MT), so it should matter most where MT is weakest. As
MT quality on high-resource pairs such as English--German~\citep{chatterjee-etal-2018-findings,chatterjee-etal-2020-findings} has
improved, APE research has shifted towards low-resource languages
(LRLs); recent WMT shared tasks
target low-resource pairs such as English--Marathi~\citep{bhattacharyya-etal-2022-findings,bhattacharyya-etal-2023-findings,zerva-etal-2024-findings}. Yet, low-resource APE has not delivered: in WMT 2023, no submitted
system improved over the English--Marathi MT baseline~\citep{bhattacharyya-etal-2023-findings}.

One possible source of that shortfall is the post-edit training signal: professional
translators interleave error correction with stylistic refinement,
mixing mandatory and discretionary changes into an inconsistent
training signal~\citep{chollampatt-etal-2020-automatic,tebbifakhr-etal-2019-effort}.
High-resource APE addresses this heterogeneity through synthetic-data
scaling, curriculum pre-training, and word-level quality-estimation (QE) supervision~\citep{deoghare-bhattacharyya-2022-iit,junczys-dowmunt-grundkiewicz-2018-ms}. Resources
for such techniques are absent for most LRLs~\citep{ranasinghe-etal-2025-sinhala,deoghare-etal-2024-together}. As we show
(Section~\ref{sec:why_collapse}), even
partial substitutes (fine-tuned QE, frontier-LLM probes,
heterogeneity filtering) do not resolve this failure. Therefore, a practitioner whose fine-tuned
checkpoint fails to beat MT cannot tell why: the model may carry a
correction signal that more work could expose, or the data may not
support APE at all. These call for opposite responses; the default
one, building more infrastructure, assumes the first case even when
the second may hold.

We introduce a black-box, inference-time diagnostic that tells these
two cases apart, without further training or annotation. Where
QE-guided APE uses external token-level annotations to choose which
tokens to keep~\citep{deoghare-etal-2025-giving}, we ask whether the
model's own per-token confidence carries the same information. The
diagnostic varies an edit-distance penalty $\lambda$ that, as it
grows, pushes the APE model from free editing towards copying the MT,
and a single sweep across $\lambda$ yields two signals. The first is
the shape of the TER (Translation Edit Rate)-vs-$\lambda$ curve: an
edit is \emph{net-positive} if it lowers TER against the reference,
so the curve is U-shaped if such edits survived training and
monotonically decreasing if none did. The second is the ordering of
constraint variants that trust the model's confidence to
increasing degrees (static, probability-scaled, entropy-scaled): if
confidence identifies net-positive edits, the confidence-aware
variants win; if it does not, the ordering reverses, and that
reversal is itself the diagnostic. Together the two signals sort a
checkpoint into one of four outcomes (Section~\ref{sec:action}): two
let the practitioner deploy immediately, one of them with a static
constraint that gives a free inference-time accuracy gain; the other
two call for a more consistent training signal before retraining.

We instantiate the diagnostic on English--Sinhala APE across
Gemma~3 1B and 4B (decoder-only) and NLLB-600M (encoder--decoder),
and validate it two ways: a contrastive low-quality-MT condition
where APE should succeed, and cross-lingual replication on
English--Marathi~\citep{bhattacharyya-etal-2022-findings} and
English--Tamil. The diagnostic exposes two failure modes, which we
hypothesize share a common cause in the heterogeneous post-edit
signal: \textbf{Binary
Collapse}, where the model copies the MT or makes an off-target edit
and loses nuanced, targeted editing; and \textbf{Confident
Miscalibration}, where high token-level confidence does not separate
unnecessary edits from useful ones.

In summary, we contribute:
\begin{itemize}
    \item A reusable, black-box, inference-time diagnostic built from
    two annotation-free signals, curve shape and constraint ordering
    (Section~\ref{sec:constrained}), with practical guidelines for
    acting on its output (Section~\ref{sec:action}).
    \item An empirical characterization of LRL APE failure under
    heterogeneous supervision, across three architectures and three
    language pairs (Section~\ref{sec:results}).
    \item The first English--Sinhala (${\sim}$66k) and a new
    English--Tamil (${\sim}$39k) \hf{https://huggingface.co/datasets/isuruwijesiri/confident-but-wrong}{APE datasets}, released with all \gh{https://github.com/IsuruMaduranga/confident-but-wrong}{code}.
\end{itemize}

\section{Related Work}
\label{sec:related}

\paragraph{APE under high- and low-resource conditions.}
High-resource APE is built on infrastructure that low-resource
pairs lack: dual-encoder Transformers~\citep{junczys-dowmunt-grundkiewicz-2018-ms}, curriculum pre-training on
millions of synthetic post-edits~\citep{deoghare-bhattacharyya-2022-iit}, and
word-level QE supervision~\citep{deoghare-etal-2023-quality}, which
together overcome the heterogeneity of professional post-edits.

The APE shared task moved from
English--German~\citep{chatterjee-etal-2018-findings,chatterjee-etal-2020-findings}
to English--Marathi~\citep{bhattacharyya-etal-2022-findings,bhattacharyya-etal-2023-findings}
and, in 2024, to English--Hindi and English--Tamil within a
quality-informed APE sub-task~\citep{zerva-etal-2024-findings}.
LLM-based APE has emerged in parallel, through few-shot prompting~\citep{raunak-etal-2023-leveraging} and error-guided prompting~\citep{ki-carpuat-2024-guiding}, but mostly on high-resource pairs;
we probe its low-resource behavior in
Appendix~\ref{sec:appendix_prompts}. For most low-resource pairs,
dedicated APE resources remain absent~\citep{ranasinghe-etal-2025-sinhala,deoghare-etal-2024-together}, and the
failure modes of fine-tuned APE under the resulting heterogeneous
post-edits remain unexamined.

\paragraph{Confidence as an edit-quality signal.}
The most directly related approach is~\citet{deoghare-etal-2025-giving}, who use external token-level QE
annotations inside Grid Beam Search to preserve correct tokens
during decoding; this requires word-level QE labels that are
unavailable for most low-resource pairs, including
English--Sinhala. Whether a model's own confidence can substitute
for such external signals is unclear, and prior work on LLM
uncertainty urges caution: networks grow overconfident after
fine-tuning~\citep{Guo2017Calibration}, token probabilities are
poorly calibrated after RLHF~\citep{kadavath2022languagemodelsmostlyknow}, and
training on model-generated data causes diversity collapse~\citep{Shumailov2024ModelCollapse}. Biasing APE towards the MT is
itself an established modeling device: \citet{huang-etal-2019-learning-copy}
add a learned copy mechanism, and~\citet{lopes-etal-2019-unbabels}
apply a conservativeness penalty at decoding time, both to improve
output quality. We use the same lever for a different purpose: swept
from free editing to forced copying, the edit-distance penalty becomes
a probe that reads whether edits are learnable and whether confidence
tracks their quality. Constrained decoding more broadly has been
explored for lexical and format constraints~\citep{post-vilar-2018-fast,hokamp-liu-2017-lexically,hu-etal-2019-improved};
repurposing it to compare confidence-trust variants as a
calibration test is, to our knowledge, new.

\section{Dataset}
\label{sec:data}
\subsection{Dataset Creation}
Our APE data comes from the \texttt{EnSiTa} English--Sinhala--Tamil
dataset~\citep{ensita2026}, covering two under-resourced languages: in the categorization of~\citet{ranathunga-de-silva-2022-languages}, Sinhala is class 2 (low-resource) and Tamil class 3 (mid-resource), and Sinhala has the less mature NLP ecosystem~\citep{Silva2019SurveyOP}. The dataset was produced by professionally post-editing machine-translated or web-mined parallel corpora across several domains, from which we recover the (source, MT, post-edit) triplets that APE requires. EnSiTa itself
releases only the final parallel corpus. We keep only triplets whose
source is identical for the MT and the post-edit, discarding items
whose source was itself edited during multi-round revision, which
breaks the fixed-source APE setup
(Appendix~\ref{sec:appendix_preprocessing}). This gives the first
${\sim}$66k English--Sinhala APE triplets across four
Google-Translate domains (D1: Literature, D2: News, D3: Wikipedia,
D4: Maths and Health) and ${\sim}$39k English--Tamil triplets across
five domains (Appendix~\ref{sec:appendix_enta}). Because EnSiTa post-editors were
\emph{permitted but not required} to make stylistic changes beyond
minimal error correction, the resulting edits are heterogeneous,
the property we analyze in Section~\ref{sec:edit_types}.

We also construct D5, a contrastive low-quality-MT
condition for this study: cleaned NLLB-corpus sources translated
with NLLB-1.3B~\citep{nllbteam2022languageleftbehindscaling}, post-edited under the same
protocol. For cross-lingual validation
(Section~\ref{sec:validation}) we use the WMT22 English--Marathi APE
data (\true{17,999} triplets;
\citealp{bhattacharyya-etal-2022-findings}).

\subsection{Edit Type Analysis: Heterogeneous, Not Consistent}
\label{sec:edit_types}

Post-edits across our domains are \textbf{heterogeneous}: no single
edit type dominates, and the mix varies by domain. Manual analysis
of 90 randomly sampled (MT, PE) pairs across D1--D3 (30 per
domain, stratified by TER bucket; Table~\ref{tab:edit_taxonomy} in
Appendix~\ref{sec:appendix_edit_taxonomy}) shows the pattern:
D1 (Literature) contains a mix of register shifts, loanword
replacements, and structural reordering with no dominant category;
D2 (News) is dominated by mechanical fixes (punctuation, spelling);
D3 (Wikipedia) mixes style edits with fluency fixes such as
transliteration of proper nouns.

We scale this analysis to \true{200} samples per domain using
GPT-5.2~\citep{openai2025gpt5} with a simplified Multidimensional
Quality Metrics (MQM-Core) taxonomy~\citep{lommel-etal-2014-mqm,freitag-etal-2021-experts}
(Section~\ref{sec:binary_collapse},
Appendix~\ref{sec:appendix_edit_classification}). On D1,
\true{49.5\%} of post-edits are purely stylistic. Three native
Sinhala speakers independently validate the classifier on
\true{100} stratified D1 samples, each judging only the single
primary category per sample, so inter-annotator agreement is
measured by Fleiss' $\kappa{=}\true{0.70}$ (per-annotator
breakdowns and confusion matrices in
Appendix~\ref{sec:appendix_mqm_validation}). D1's post-edits are the most heterogeneous of our domains; D2 and D3
are progressively more consistent.

\section{The Investigation}
\label{sec:methods}

\subsection{Task Formulation}
We first fix notation for the APE task and the quantities the
diagnostic reads. APE takes a source sentence
$x^{\text{src}}=(x^{\text{src}}_1,\dots,x^{\text{src}}_N)$ and its
machine translation
$x^{\text{mt}}=(x^{\text{mt}}_1,\dots,x^{\text{mt}}_M)$, and produces
a post-edit $\mathbf{y}=(y_1,\dots,y_T)$. The post-edit
is evaluated against a human reference $\mathbf{y}^\star$ using
$\mathrm{TER}(\mathbf{y},\mathbf{y}^\star)$. At each step $t$, the
fine-tuned APE model $M$ produces a distribution
$p_t(\cdot)=p(\cdot \mid x^{\text{src}}, x^{\text{mt}}, y_{<t})$
over vocabulary $\mathcal{V}$. Throughout, $x^{\text{mt}}_j$
denotes the $j$-th MT token, and
$\mathcal{W}_t \subseteq \{1,\dots,M\}$
denotes the indices of a local alignment window of MT tokens
centered on the current alignment pointer (defined in
Section~\ref{sec:constrained}). Recall that an edit is \emph{net-positive} if it reduces TER against the
reference.

\subsection{Models and Training}
\label{sec:models}

We first train each model on the APE triplets with standard supervised
fine-tuning (\emph{vanilla fine-tuning}), then apply constrained-decoding
variants to the same checkpoints at inference time.

We evaluate the following model families: \textbf{Gemma~3 1B-IT and 4B-IT} \citep{Gemma3},
instruction-tuned decoder-only LLMs with LoRA/QLoRA fine-tuning,
and \textbf{NLLB-600M} \citep{nllbteam2022languageleftbehindscaling}, a multilingual
encoder--decoder NMT model, fine-tuned
with LoRA. NLLB substitutes for the standard dual-encoder APE
pipeline, which is infeasible for English--Sinhala. The available encoders~\cite{dhananjaya-etal-2022-bertifying} are not strong enough and large parallel corpora are absent~\citep{ranasinghe-etal-2025-sinhala}.

We omit synthetic data, which would mask the heterogeneous nature
of real post-edits. All models are
trained with three independent random seeds (42, 52, 62); we
report mean~$\pm$~std throughout. Full hyperparameters and the input formats for both families are
in Appendix Table~\ref{tab:training_details}; evaluation metric
details in Appendix~\ref{sec:appendix_metrics}.

\begin{table}[t!]
  \centering
  \resizebox{\columnwidth}{!}{%
  \begin{tabular}{lrrrrr}
  \toprule
  & \textbf{D1} & \textbf{D2} & \textbf{D3} & \textbf{D4} & \textbf{D5} \\
  & \textbf{Lit.} & \textbf{News} & \textbf{Wiki} & \textbf{Math/H} & \textbf{NLLB} \\
  \midrule
  Train & \true{5,728} & \true{13,305} & \true{17,471} & \true{5,754} & \true{10,332} \\
  Valid & \true{716} & \true{1,663} & \true{2,184} & \true{719} & \true{1,291} \\
  Test & \true{716} & \true{1,664} & \true{2,184} & \true{720} & \true{1,292} \\
  \midrule
  Total & \true{7,160} & \true{16,632} & \true{21,839} & \true{7,193} & \true{12,915} \\
  MT TER & \true{24.98} & \true{7.93} & \true{14.89} & \true{24.30} & \true{59.19} \\
  \midrule
  MT source & \multicolumn{4}{c}{Google Translate} & NLLB-1.3B \\
  \bottomrule
  \end{tabular}}
  \caption{Experimental splits D1--D5 (${\sim}$\true{66k} triplets). D1--D4 use Google Translate MT; D5 is a contrastive low-quality NLLB-1.3B condition. MT TER is test-set TER of unedited MT vs.\ PE. Full statistics in Appendix Table~\ref{tab:ensi_dataset}.}
  \label{tab:dataset}
  \end{table}

\subsection{Binary Collapse: Copy or Off-Target Edit}
\label{sec:binary_collapse}

On D1 (Literature), vanilla fine-tuning \emph{degrades} TER across
all three model families (Table~\ref{tab:cross_model}): Gemma 1B
by \true{0.87} points, 4B by \true{0.43}, and NLLB-600M by
\true{5.80}. Within the decoder-only family, scale mitigates the
degradation; the encoder--decoder NLLB-600M fails most severely.

The model's editing behavior has two modes. It copies the MT
verbatim on most sentences (\true{71.5\%} of the \true{200} classified
D1 samples, every one of which requires an edit;
Table~\ref{tab:edit_decomposition}). On the minority of sentences
where it does edit, the edit rarely reaches the human correction (only \true{4.0\%} of outputs
match PE). We term this \textbf{Binary Collapse}: fine-tuning collapses
nuanced, targeted editing into copying or off-target edits. The collapse
is one of edit \emph{quality}, not magnitude (the edits are moderate,
not wholesale rewrites), and it is specific to the inconsistent D1
signal: on D2, D3, and the contrastive D5 condition the model's edits
are net-positive, whereas on D1 they are net-negative, leaving the
fine-tuned model below the MT baseline
(Appendix~\ref{sec:appendix_edit_classification} Table~\ref{tab:edit_behavior}).

Independent confirmation comes from a Chain-of-Thought
self-tagging experiment (Appendix~\ref{sec:appendix_failed}), in
which the model is trained to predict its own edit level before
post-editing. Its predicted distribution collapses the minor class
from \true{20.9\%} in the gold labels to \true{0.1\%}
(Table~\ref{tab:class_collapse}), with \true{90\%} of gold-minor
samples tagged as \texttt{[major]}: the model has internalized a
binary none-or-major view of editing. This self-assessment failure is
also an early sign of miscalibration, the model staying confident in
edit-level predictions that are wrong; we establish that miscalibration
directly in Section~\ref{sec:token_confidence}. It is consistent with
evidence that fine-tuned LLMs lose reliable self-knowledge under
distribution shift~\citep{kadavath2022languagemodelsmostlyknow}.

Two other training-time interventions, Hidden State Anchoring
(HSA) and Edit-Level Conditioning (ELC), also fail to reduce
over-editing (Appendix Table~\ref{tab:failed}). These failures led us to hypothesize \textbf{Confident
Miscalibration}: the model's own confidence cannot tell good edits
from bad.

\subsection{Constrained Decoding as a Diagnostic}
\label{sec:constrained}

We test the Confident Miscalibration hypothesis with constrained
decoding, an inference-time diagnostic that needs no retraining. It simply
reads two signals, one per failure mode: the shape of the
TER-vs-$\lambda$ curve reads whether the model learned useful edits
(Binary Collapse), and the ordering of the constraint variants reads
whether the model's confidence can find those edits (Confident
Miscalibration). The two modes share one suspected cause and one
remedy (Section~\ref{sec:why_collapse}); we keep them apart because
they prescribe different actions (Section~\ref{sec:action}).

The first signal is the curve shape. Raising $\lambda$
from~0 (unconstrained) toward large values forces the model to copy,
meaning that at large $\lambda$, TER equals the MT baseline.
Read against that baseline, the TER-vs-$\lambda$ curve reveals whether
the model learned any net-positive edits in training: a U-shaped
curve dips below it at some intermediate $\lambda$, where partial
editing beats copying, so it learned some. A monotonically decreasing curve never falls below the
baseline, reaching it only as editing is suppressed, so every model
edit is, on average, harmful.

All variants penalize divergence from the MT output using
\emph{window-based alignment}: a pointer tracks position in the
MT sequence, and a window $\mathcal{W}_t$ of $2k{+}1$ MT positions
centered on the current pointer defines the set of MT tokens
allowed at step~$t$. The window distance is binary:
$d(y_t, \mathcal{W}_t) = 0$ if $y_t = x^{\text{mt}}_j$ for some
$j \in \mathcal{W}_t$, else $1$.

\paragraph{Static ($\lambda_{\text{static}}$).}
A uniform penalty $\lambda \ge 0$ is applied at every step,
\textbf{ignoring model confidence}:
\begin{equation}
    \hat{p}(y_t) \propto p_t(y_t) \cdot \exp\bigl(-\lambda \cdot d(y_t, \mathcal{W}_t)\bigr)
\label{eq:static}
\end{equation}

\paragraph{Probability-scaled (PMT; $\lambda_{\text{pmt}}$).}
The penalty is scaled by how much probability mass the model
puts on MT-window tokens:
\begin{equation}
    \lambda_t = \lambda \cdot P_{\text{copy}}^\beta
\label{eq:prob}
\end{equation}
where $P_{\text{copy}} = \max_{j \in \mathcal{W}_t} p_t(x^{\text{mt}}_j)$
is the maximum probability the model assigns to any MT-window
token, and $\beta > 0$ controls sharpness. When the model already
prefers an MT-window token ($P_{\text{copy}}$ large), the
constraint is reinforced; when it does not ($P_{\text{copy}}$
small), the constraint relaxes.

\paragraph{Entropy-scaled ($\lambda_{\text{entropy}}$).}
The penalty is scaled by how uncertain the model is at step $t$:
\begin{equation}
    \lambda_t = \lambda \cdot \left(\frac{H(p_t)}{H_{\max}}\right)^\gamma
\label{eq:entropy}
\end{equation}
where $H(p_t) = -\sum_v p_t(v)\log p_t(v)$ is the entropy of the
output distribution at step $t$, $H_{\max} = \log|\mathcal{V}|$ is
the maximum possible entropy, and $\gamma > 0$ controls how
sharply the constraint switches off as the model becomes
confident.

The three variants form a hierarchy that trusts the model's
confidence to increasing degrees: static ignores it,
probability-scaled relaxes the constraint where the model prefers
to diverge from the MT, and entropy-scaled relaxes wherever the model
is confident. Their ordering is the second signal. If the model's
confidence correlates with which of its edits are net-positive, the
confidence-aware variants should win; if it does not, the ordering
reverses to Static $>$ Probability $>$ Entropy, the signature of
Confident Miscalibration, confirmed in
Section~\ref{sec:miscalibration}. A fourth variant,
sequence-scaled ($\lambda_{\text{seq-NLL}}$), replaces the
per-token signal with the running confidence over the whole
generated prefix (Appendix~\ref{sec:appendix_seqnll}); it checks
that the ordering is not an artifact of token-level signals.
Search ranges and the validation-selected values for
$\lambda$, $\gamma$, $\beta$, and the window size $k$ are reported in
Table~\ref{tab:hyperparams}; Appendix~\ref{sec:appendix_ksens}
shows the diagnostic is stable across $k$.

\section{Results and Analysis}
\label{sec:results}

This section follows the diagnostic's two signals.
Sections~\ref{sec:lambda_curve} and~\ref{sec:miscalibration} read
the curve shape and the constraint ordering on D1,
Section~\ref{sec:token_confidence} confirms both at the token
level, and Sections~\ref{sec:generalization}
to~\ref{sec:crosslingual} replicate the pattern across
architectures, MT quality, and language pairs.

\subsection{No Net-Positive Edits on D1}
\label{sec:lambda_curve}

\begin{figure*}[t!]
  \centering
  \includegraphics[width=\textwidth]{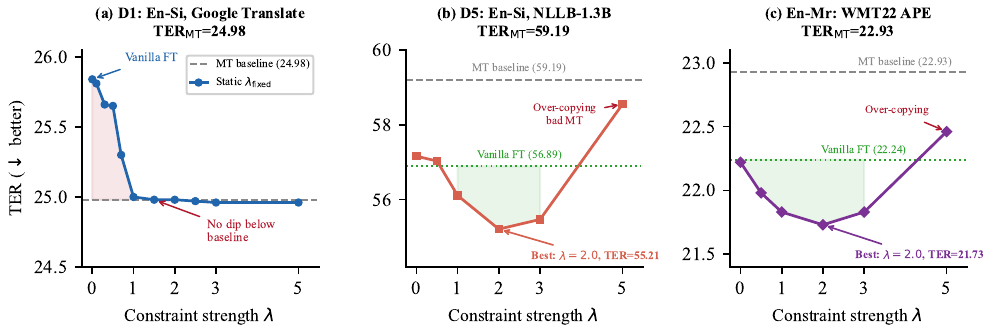}
  \caption{TER vs.\ $\lambda$ curves (Gemma 3 1B, seed 42). \textbf{Left:} D1 (TER~\true{24.98}): monotonic. \textbf{Center:} D5 (TER~\true{59.19}): U-shaped, min $\lambda{=}2.0$. \textbf{Right:} En-Mr (WMT22, TER~\true{22.93}): U-shaped, min $\lambda{=}2.0$. En-Mr's MT quality matches D1 yet its curve is U-shaped: the pattern is consistent with a post-edit-distribution effect, not an MT-quality effect.}
  \label{fig:lambda_curves}
\end{figure*}

Figure~\ref{fig:lambda_curves} (left) plots TER as a function of $\lambda$ for the static constraint on D1 (seed~42; full sweep in Appendix Table~\ref{tab:lambda_sweep}). We observe a \textbf{monotonically decreasing} curve from TER~\true{25.84} ($\lambda{=}0$) toward \true{24.97} ($\lambda{\geq}2.5$), with no dip below the MT baseline at any $\lambda$. This indicates that every deviation from MT is, on average, harmful: the model has not learned a net-positive editing pattern, and copying the MT outright is the best it can do. The curve shape is not an artifact of TER: recomputing the sweep under chrF++ and BLEU preserves the monotonic-versus-U-shaped distinction on D1, D5, and English--Marathi (Appendix~\ref{sec:appendix_metric_robustness}).

\subsection{Confident Miscalibration: Confidence Does Not Identify Net-Positive Edits}
\label{sec:miscalibration}

Binary Collapse (Section~\ref{sec:binary_collapse}) tells us \emph{what} the model does wrong. The constraint ordering tells us \emph{why}. Table~\ref{tab:constrained} gives the central result: the static constraint, which ignores the model's confidence, beats every confidence-aware variant, so per-token confidence does not separate net-positive edits from unnecessary ones.

\begin{table}[t!]
\centering
\small
\setlength{\tabcolsep}{3.5pt}
\begin{tabular}{llcc}
\toprule
\textbf{Method} & \textbf{Logic} & \textbf{TER} $\downarrow$ & \textbf{$\Delta$TER} \\
\midrule
MT Baseline & -- & \true{24.98} & -- \\
Vanilla FT & -- & \true{25.85\scriptsize{$\pm$0.02}} & \true{$-$0.87} \\
\midrule
$\lambda_{\text{entropy}}$ & Trust if confident & \true{25.72\scriptsize{$\pm$0.09}} & \true{$-$0.74} \\
$\lambda_{\text{pmt}}$ & Trust if hates MT & \true{25.37\scriptsize{$\pm$0.25}} & \true{$-$0.39} \\
$\lambda_{\text{seq-NLL}}$ & Trust if seq.\ confident & \true{25.32\scriptsize{$\pm$0.21}} & \true{$-$0.34} \\
\rowcolor{green!10}
$\lambda_{\text{static}}$ & \textbf{Don't trust model} & \textbf{\true{24.98\scriptsize{$\pm$0.08}}} & \textbf{\true{0.00}} \\
\bottomrule
\end{tabular}
\caption{Constrained decoding on D1 (Gemma 3 1B, mean $\pm$ std across 3 seeds). The \textbf{Static $>$ Probability $>$ Entropy} ordering is the signature of Confident Miscalibration (zero-shot: Appendix~\ref{sec:appendix_prompts}, Table~\ref{tab:gpt5_results}). The sequence-level variant patterns with the token-level ones, so the ordering is not an artifact of token-level signals.}
\label{tab:constrained}
\end{table}

The ordering confirms Confident Miscalibration. The static constraint ($\lambda{=}\true{2.5}$) recovers the MT baseline exactly, where the $\lambda$ curve also flattened (Section~\ref{sec:lambda_curve}). The probability variant lands in the middle, since $P_{\text{copy}}$ carries some information. The entropy variant is worst: its TER stays flat across all base $\lambda$ (Table~\ref{tab:lambda_sweep}). The sequence-level variant, which sets the penalty from the model's confidence over the whole generated prefix rather than a single token, lands with the token-level variants (TER \true{25.32$\pm$0.21}). No confidence signal we test, token-level or sequence-level, reaches the confidence-free static baseline, so the failure lies in the confidence itself, not in how we integrate it.

\subsection{Token-Level Confidence: Direct Evidence for Both Failures}
\label{sec:token_confidence}
We measure model confidence directly on correct vs.\ incorrect editing decisions. We classify each of the \true{20,784} tokens generated by the vanilla 1B model (seed~42, D1 test, window $k{=}2$) by whether it matches the MT and the \emph{gold PE} (the human reference post-edit $\mathbf{y}^\star$). This gives four categories: \emph{correct copy} (matches both), \emph{correct edit} (differs from MT, matches gold PE), \emph{unnecessary edit} (differs from MT, but MT was correct), and \emph{missed edit} (copies MT, but gold PE differs). Table~\ref{tab:confidence} presents the results.

\begin{table}[t!]
\centering
\small
\begin{tabular}{lccc}
\toprule
\textbf{Token Category} & \textbf{\%} & \textbf{Mean $p$} & \textbf{Mean $H$} \\
\midrule
Correct copy & \true{64.5} & \true{0.95} & \true{0.30} \\
Missed edit & \true{29.0} & \true{0.95} & \true{0.33} \\
\rowcolor{red!8}
Unnecessary edit & \true{5.2} & \true{0.80} & \true{0.91} \\
Correct edit & \true{1.2} & \true{0.81} & \true{0.84} \\
\bottomrule
\end{tabular}
\caption{Token-level confidence for Gemma 3 1B on D1 (vanilla decoding). $p$: mean token probability; $H$: mean output entropy.}
\label{tab:confidence}
\end{table}

At the token level, both failure modes carry a confidence signature. Binary Collapse: the model copies \true{93.5\%} of tokens, and \emph{missed edits} ($p{=}\true{0.95}$, $H{=}\true{0.33}$) are confidence-indistinguishable from \emph{correct copies} ($p{=}\true{0.95}$, $H{=}\true{0.30}$), so it cannot tell which tokens need editing. Confident Miscalibration: when the model does edit, \emph{unnecessary edits} ($p{=}\true{0.80}$, $H{=}\true{0.91}$) and \emph{correct edits} ($p{=}\true{0.81}$, $H{=}\true{0.84}$) are equally confident, so it cannot tell which edits are useful. The overlap is starkest at the high-confidence end: \true{59.4\%} of unnecessary edits and \true{54.8\%} of correct edits both fall above $p{>}0.9$, as do \true{84.8\%} of missed edits (full distribution in Appendix~\ref{sec:appendix_results} Figure~\ref{fig:confidence_scatter}).

\begin{table}[t!]
\centering
\small
\setlength{\tabcolsep}{4pt}
\begin{tabular}{lccc}
\toprule
\textbf{Token Category} & \textbf{D1} & \textbf{D2} & \textbf{D5} \\
\midrule
Correct copy & \true{64.5} (\true{.95}) & \true{87.0} (\true{.99}) & \true{29.4} (\true{.90}) \\
Missed edit & \true{29.0} (\true{.95}) & \true{7.6} (\true{.97}) & \true{36.9} (\true{.86}) \\
Unnecessary edit & \true{5.2} (\true{.80}) & \true{3.7} (\true{.90}) & \true{28.4} (\true{.77}) \\
Correct edit & \true{1.2} (\true{.81}) & \true{1.7} (\true{.95}) & \true{5.3} (\true{.84}) \\
\bottomrule
\end{tabular}
\caption{The Table~\ref{tab:confidence} breakdown repeated on D2 (clean MT) and the U-shaped D5 (Gemma 3 1B, seed 42). Each cell gives the token share in \% (mean $p$ in parentheses).}
\label{tab:confidence_cross}
\end{table}

The breakdown is not specific to the failing D1 setting. Repeating it on D2 (clean MT) and on the U-shaped D5 (Table~\ref{tab:confidence_cross}) shows what changes as the post-edits become more consistent, and what does not. The share of correct edits rises (\true{1.2\%} to \true{1.7\%} to \true{5.3\%}): the model does make useful edits when the data supports them, consistent with the D1 data contributing to the failure. The confidence overlap, however, persists everywhere: in every setting \true{47\%} to \true{77\%} of unnecessary edits still carry probability above \true{0.9}, and unnecessary edits outnumber correct ones by two to five times. Even when APE works, the model's own confidence is not a reliable gate.

The entropy variant defers to these high-confidence predictions, so it fails. Neither structured prompting nor recalibration resolves the failure. GPT-5.2 with explicit Stage-1 gating reaches only TER~\true{36.75} (gating precision~\true{0.75}, \true{2} points above an always-edit baseline; Appendix~\ref{sec:appendix_two_stage}). Temperature scaling at $T^*{=}\true{2.0}$ worsens unconstrained generation while preserving the ordering (Appendix~\ref{sec:appendix_temperature}).

\needspace{5\baselineskip}
\subsection{The Same Failure Across Architectures, Scale, and Domains}
\label{sec:generalization}

Table~\ref{tab:cross_model} compares all three model families with constrained decoding on D1.

\begin{table}[t!]
\centering
\small
\setlength{\tabcolsep}{4pt}
\begin{tabular}{llccc}
\toprule
\textbf{Model} & \textbf{Method} & \textbf{TER} $\downarrow$ & \textbf{BLEU} $\uparrow$ & \textbf{$\Delta$TER} \\
\midrule
\multicolumn{5}{l}{\textit{Gemma 3 1B (decoder-only, LoRA):}} \\
& Base (zero-shot) & \true{190.31} & \true{8.88} & \true{$-$165.33} \\
& Vanilla FT & \true{25.85} & \true{67.16} & \true{$-$0.87} \\
& $\lambda_{\text{entropy}}$ & \true{25.72} & \true{67.23} & \true{$-$0.74} \\
& $\lambda_{\text{pmt}}$ & \true{25.37} & \true{67.45} & \true{$-$0.39} \\
& $\lambda_{\text{static}}$ & \textbf{\true{24.98}} & \textbf{\true{67.38}} & \textbf{\true{0.00}} \\
\midrule
\multicolumn{5}{l}{\textit{Gemma 3 4B (decoder-only, QLoRA):}} \\
& Base (zero-shot) & \true{93.81} & \true{24.32} & \true{$-$68.83} \\
& Vanilla FT & \true{25.41} & \true{67.37} & \true{$-$0.43} \\
& $\lambda_{\text{entropy}}$ & \true{25.05} & \true{67.53} & \true{$-$0.07} \\
& $\lambda_{\text{pmt}}$ & \true{24.90} & \true{67.76} & \true{+0.08} \\
& $\lambda_{\text{static}}$ & \textbf{\true{24.83}} & \textbf{\true{67.52}} & \textbf{\true{+0.15}} \\
\midrule
\multicolumn{5}{l}{\textit{NLLB-600M (encoder--decoder, LoRA):}} \\
& Vanilla FT & \true{30.78} & \true{61.98} & \true{$-$5.80} \\
& $\lambda_{\text{entropy}}$ & \true{31.28} & \true{61.57} & \true{$-$6.30} \\
& $\lambda_{\text{pmt}}$ & \true{29.97} & \true{62.58} & \true{$-$4.98} \\
& $\lambda_{\text{static}}$ & \textbf{\true{29.59}} & \textbf{\true{62.63}} & \textbf{\true{$-$4.61}} \\
\bottomrule
\end{tabular}
\caption{Cross-model constrained decoding on D1. \textbf{All three architectures show the same ordering: Static $>$ Probability $>$ Entropy.}}
\label{tab:cross_model}
\end{table}

\paragraph{Scale mitigates but does not fix.} Gemma 4B halves the degradation (\true{$-$0.43} vs.\ \true{$-$0.87} for 1B), and is the only model to \emph{beat} the MT baseline with static constraint ($\Delta$TER${=}\true{+0.15}$). Its $\lambda$ curve is weakly U-shaped (Figure~\ref{fig:lambda_scale} in Appendix~\ref{sec:appendix_cross_model}), unlike 1B's monotonic curve. Yet the Static~$>$~PMT~$>$~Entropy ordering still holds, with entropy TER essentially flat (\true{25.05}--\true{25.08}). Confidence remains uninformative even with 4$\times$ more parameters: scale mitigates Binary Collapse but not Confident Miscalibration.

\paragraph{Encoder-decoder failure.} NLLB-600M degrades by \true{$-$5.80} TER, far worse than Gemma 1B (\true{$-$0.87}) and 4B (\true{$-$0.43}). Even aggressive static constraint only reaches TER~\true{29.59} ($\lambda{=}5.0$), still \true{4.61} above MT. A plateau at $\lambda{\geq}10$ confirms constraint saturation (Figure~\ref{fig:lambda_scale}): the over-editing is too severe for inference-time correction. However, this failure is task-specific: the same model succeeds on D5 ($\Delta$TER${=}\true{+2.05}$; Section~\ref{sec:validation}), showing that the architecture can succeed under a contrasting condition.

\paragraph{Domain robustness.} Vanilla fine-tuning degrades TER on D1 and D4 but not D2 (News) or D3 (Wiki; Appendix Table~\ref{tab:domain}). Qualitative examples of recurring failure modes (unnecessary edits on correct MT, content corruption, hallucinated corrections) are in Appendix~\ref{sec:appendix_qualitative}.

\subsection{Contrastive Validation}
\label{sec:validation}

A valid diagnostic must also detect success. D5 is such an example. Its MT is far worse than D1's (TER~\true{59.19} vs.\ \true{24.98}), so post-editors mostly fix real errors instead of making optional style changes. That gives D5 a consistent, learnable signal, unlike D1. The same models that fail on D1 succeed on D5 (Table~\ref{tab:d5_results}): vanilla fine-tuning \emph{improves} TER by \true{+2.30} for Gemma 1B and \true{+2.05} for NLLB-600M, reversing their \true{$-$0.87} and \true{$-$5.80} on D1. So the D1 failure comes from the post-edit distribution, not the models or fine-tuning.

\begin{table}[t!]
\centering
\small
\begin{tabular}{llccc}
\toprule
& & \textbf{TER} $\downarrow$ & \textbf{BLEU} $\uparrow$ & \textbf{$\Delta$TER} \\
\midrule
\multicolumn{5}{l}{\textit{D1: High-quality MT (Google Translate):}} \\
& MT Baseline & \true{24.98} & \true{67.38} & -- \\
& Vanilla FT & \true{25.85} & \true{67.16} & \true{$-$0.87} \\
& $\lambda_{\text{static}}$ & \true{24.98} & \true{67.38} & \true{0.00} \\
\midrule
\multicolumn{5}{l}{\textit{D5: Low-quality MT (NLLB-1.3B), Gemma 1B:}} \\
& MT Baseline & \true{59.19} & \true{30.54} & -- \\
& Vanilla FT & \true{56.89} & \true{31.60} & \true{+2.30} \\
& $\lambda_{\text{static}}$ & \textbf{\true{55.21}} & \textbf{\true{33.12}} & \textbf{\true{+3.98}} \\
& $\lambda_{\text{pmt}}$ & \true{56.58} & \true{32.20} & \true{+2.61} \\
& $\lambda_{\text{entropy}}$ & \true{56.99} & \true{31.88} & \true{+2.20} \\
\midrule
\multicolumn{5}{l}{\textit{D5: Low-quality MT (NLLB-1.3B), NLLB-600M:}} \\
& Vanilla FT & \true{57.14} & \true{32.01} & \true{+2.05} \\
\bottomrule
\end{tabular}
\caption{D1 vs.\ D5 (seed 42). On D5 (genuine error correction) both Gemma 1B and NLLB-600M improve over MT; constrained decoding further improves Gemma 1B, and Static~$>$~PMT~$>$~Entropy still holds.}
\label{tab:d5_results}
\end{table}

The Gemma 1B $\lambda$ curve for D5 is \textbf{U-shaped} (Figure~\ref{fig:lambda_curves}, center): TER reaches a minimum of \true{55.21} at $\lambda{=}\true{2.0}$, well below both vanilla FT and the MT baseline, then rises as the constraint forces copying of low-quality MT. NLLB-600M has its minimum at $\lambda{\approx}0$ (Appendix Figure~\ref{fig:lambda_scale}): on D5, its TER is minimized without a constraint.

\subsection{Cross-Lingual Replication}
\label{sec:crosslingual}

So far, the D1 failure and the D5 success both come from
English--Sinhala. We replicate the diagnostic on two more language
pairs: English--Marathi (Indo-Aryan, like Sinhala) and English--Tamil
(Dravidian). In both, the two diagnostic signals, curve shape and
constraint ordering, behave just as they do on English--Sinhala.

On WMT22 \textbf{English--Marathi} APE (\true{17,999} triplets; \citealp{bhattacharyya-etal-2022-findings}), MT quality matches D1 (TER~\true{22.93} vs.\ \true{24.98}). Yet vanilla fine-tuning \emph{improves} TER by \true{+0.69} (Table~\ref{tab:enmr_results}). The $\lambda$ curve is U-shaped (Figure~\ref{fig:lambda_curves}, right), and Static~$>$~PMT~$>$~Entropy holds.

\textbf{English--Tamil} contributes a new News corpus (\true{10,000} triplets, Google Translate MT, TER$_{\mathrm{MT}}{=}\true{20.57}$). Vanilla fine-tuning degrades TER by \true{$-$1.21} (Table~\ref{tab:enta_results}), and the $\lambda$ curve is monotonic (Figure~\ref{fig:enta_contrastive}a), mirroring D1. Tamil News shares D1's heterogeneous edit profile (\true{44}/\true{20}/\true{36} none/minor/major; Appendix Table~\ref{tab:enta_dataset}). A within-language contrastive condition using NLLB-1.3B MT (TER$_{\mathrm{MT}}{=}\true{47.54}$) on the same sources produces a U-shaped curve (Figure~\ref{fig:enta_contrastive}b, minimum at $\lambda{=}\true{3.0}$), mirroring the D1/D5 pattern.

\begin{table}[t!]
  \centering
  \small
  \begin{tabular}{lccc}
  \toprule
  \textbf{System} & \textbf{TER} $\downarrow$ & \textbf{BLEU} $\uparrow$ & \textbf{$\Delta$TER} \\
  \midrule
  MT Baseline & \true{22.93} & \true{67.01} & -- \\
  Vanilla FT & \true{22.24} & \true{69.81} & \true{+0.69} \\
  \midrule
  $\lambda_{\text{entropy}}$ & \true{22.06} & \true{69.95} & \true{+0.87} \\
  $\lambda_{\text{pmt}}$ & \true{21.83} & \true{69.96} & \true{+1.10} \\
  \rowcolor{green!10}
  $\lambda_{\text{static}}$ & \textbf{\true{21.73}} & \textbf{\true{69.72}} & \textbf{\true{+1.20}} \\
  \bottomrule
  \end{tabular}
  \caption{English--Marathi validation (Gemma 3 1B, seed 42, WMT22 APE dev set; test refs unavailable). \textbf{Static~$>$~PMT~$>$~Entropy} holds cross-lingually. Full sweeps in Appendix Table~\ref{tab:enmr_lambda}.}
  \label{tab:enmr_results}
  \end{table}

\begin{table}[t!]
\centering
\small
\begin{tabular}{lccc}
\toprule
\textbf{System} & \textbf{TER} $\downarrow$ & \textbf{BLEU} $\uparrow$ & \textbf{$\Delta$TER} \\
\midrule
MT Baseline & \true{20.57} & \true{73.17} & -- \\
Vanilla FT & \true{21.77} & \true{71.39} & \true{$-$1.21} \\
\midrule
$\lambda_{\text{static}}$ & \true{20.68} & \true{73.00} & \true{$-$0.11} \\
\bottomrule
\end{tabular}
\caption{English--Tamil validation (News, Gemma 3 1B, seed 42, static constraint $\lambda{=}5.0$). The monotonic $\lambda$ curve approaches but never crosses the MT baseline. Full sweep in Appendix Table~\ref{tab:enta_lambda}; per-domain in Table~\ref{tab:enta_domain}.}
\label{tab:enta_results}
\end{table}

\section{Discussion}
\label{sec:discussion}

\subsection{What Drives Both Failures: Inconsistency, Not Edit Type}
\label{sec:why_collapse}

Section~\ref{sec:generalization} showed the failure is domain-specific,
which points at the data, not the model. Our hypothesis, which the
evidence below supports but does not prove, is that the heterogeneous
post-edits (Section~\ref{sec:edit_types}) are the source, and that the
problem is not their variety but their \emph{inconsistency}: the same
kind of error gets fixed in conflicting ways, so no stable rule
connects a correction to its context. Under this hypothesis, the model
cannot learn a coherent editing strategy, so it falls back to copying
or off-target rewriting (Binary Collapse); and because no consistent
signal links a correction to its quality, the model's confidence cannot
tell a useful edit from a harmful one (Confident Miscalibration).
The two modes are thus symptoms of one suspected cause, sharing
one remedy; we keep them apart only because they are read from
different signals and prescribe different actions
(Section~\ref{sec:action}). External QE signals fail the same way
(Appendix~\ref{sec:appendix_qe}). The evidence here is
correlational, but three independent lines converge on it.

D3 is a useful contrast. If edit type drove the failure, two domains
with the same edit-type mix would behave alike. D3 has almost the same
mix as D1: identical \true{41.0\%} Style, and \true{48.0\%} vs
\true{49.5\%} subjective overall
(Table~\ref{tab:edit_cross_domain}). Yet on the full test set, D3's
edits are net-positive while D1's are net-negative (per-edit $\Delta$TER
\true{$-$1.1} vs \true{$+$6.3};
Appendix Table~\ref{tab:edit_behavior}). With the mix held fixed, the results are consistent with a more
learnable correction signal. Annotation policy, MT quality, and language
family are also less plausible explanations (Sections~\ref{sec:validation},~\ref{sec:crosslingual}):
D2's mechanical edits are learnable under the same policy,
English--Marathi matches D1's MT quality yet edits net-positively, and
English--Tamil reproduces the failure though it is Dravidian, not
Indo-Aryan.

\paragraph{Gradient analysis.} Inspecting the model's gradients supports the same account (Appendix~\ref{sec:appendix_gradient}): two independent instruments, one white-box and one black-box, are consistent with inconsistent post-edits as the driver. First, the gradient has
no stable target: on D1, Style is the largest edit group yet produces
the \emph{smallest} mean gradient (norm \true{1.21}), because its
subtypes pull in opposite directions and cancel, and its per-sample
loss is the most variable of any group (coefficient of variation
\true{1.11}). Second, the model cannot tell the edit types apart: on
D1, their gradients overlap (max pairwise cosine \true{0.275}), while
on D5, where the corrections are genuine, they are near-orthogonal
(\true{0.051}). Third, the same signal sharpens when its surroundings
are consistent: Fluency's mean gradient norm rises from \true{1.46} on
D1 to \true{3.60} on D5, as morphological and punctuation corrections
diluted by heterogeneous neighbors on D1 emerge as the clearest signal
on D5. The geometry tracks the $\lambda$ curve: both collapse on D1
and separate on D5.

If the hypothesis is right and inconsistency rather than any single
edit type drives the failure, removing one edit type should not
repair it.
Retraining Gemma~3 1B on D1 with all Style and Retranslation samples
removed confirms this. The TER-vs-$\lambda$ curve stays monotonically
decreasing, never crossing the MT baseline, and shows no advantage over
a size-matched random subsample (Appendix~\ref{sec:appendix_edit_classification} Table~\ref{tab:filter_lambda}). The Accuracy and Fluency edits that
remain are themselves inconsistent, matching the cross-domain pattern
(Table~\ref{tab:edit_cross_domain}) and the gradient analysis above.

\subsection{Implications for Quality Estimation and Confidence}
External QE signals do not reliably identify segments needing edits: fine-tuned XLM-R~\citep{conneau-etal-2020-unsupervised} reaches only AUROC~\true{0.745} on D1, no better when trained on test data (Appendix~\ref{sec:appendix_qe}, Table~\ref{tab:qe_results}). Pretrained QE also fails: LaBSE~\citep{feng-etal-2022-language} similarity is below chance and COMET-QE~\citep{rei-etal-2022-cometkiwi} near it (AUROC \true{0.45} and \true{0.58}, Table~\ref{tab:qe_results}). QE-guided APE pipelines should therefore check post-edit consistency before investing in QE infrastructure~\citep{deoghare-etal-2025-giving}.

The model's own confidence is no more reliable, at least across the signals we test (token-level probability and entropy, and a sequence-level one): it stays certain even when wrong. A GPT-5.2 probe across zero-shot, few-shot, RAG, and two-stage conditions fails the same way (Appendix~\ref{sec:appendix_prompts}), pointing to a task-level rather than capacity-level failure. Our validated claims are confined to APE, across three architectures and three language pairs. Because the diagnostic is black-box and not tied to APE, it may extend to other editing tasks with similarly inconsistent supervision, but that extension is untested and remains future work.

\subsection{From Diagnosis to Action}
\label{sec:action}

A single $\lambda$ sweep does more than diagnose: in the favorable case,
it also improves accuracy. It is also cheap: each $\lambda$ value costs
one decoding pass over the evaluation set, and the constraint logic
itself adds negligible overhead ($<$5\% wall-clock;
Appendix~\ref{sec:appendix_decoding}). The shape of the curve and the
location of its minimum sort each fine-tuned checkpoint we evaluate into one of four
outcomes (Appendix~\ref{sec:appendix_practitioner}), each with a
concrete next step:

\begin{enumerate}
\item \textbf{U-shaped, dipping below the MT baseline.} Net-positive
edits survived training. Deploy the static constraint at the optimal
$\lambda$ for an inference-time accuracy gain, with no retraining or
annotation (as shown robustly on D5 and English--Marathi,
Tables~\ref{tab:d5_results} and~\ref{tab:enmr_results}; marginally on
Gemma~4B at D1).
\item \textbf{Minimum at $\lambda{=}0$.} The model already edits
net-positively without any constraint. Deploy the vanilla checkpoint
unchanged (as for D2).
\item \textbf{Monotonic, reaching the MT baseline.} No net-positive
edits survived, and copying is optimal. Copy the MT for now, and make
the post-edits more consistent before training again (as for D1).
\item \textbf{Monotonic, plateauing above the MT baseline.} Over-editing is too severe to undo at inference. Fix the data or MT
quality before training again (NLLB-600M on D1).
\end{enumerate}

Whenever a constraint helps, deploy the \emph{static} variant: the
Static~$>$~PMT~$>$~Entropy ordering (Section~\ref{sec:miscalibration})
shows that trusting the model's confidence only costs accuracy
(Appendix~\ref{sec:appendix_why_static}).

Outcomes 3 and 4 cannot be repaired at inference; the fix has to come from
the data. The suspected cause is inconsistent post-edits
(Section~\ref{sec:why_collapse}), so the corrections must be made more
consistent before retraining. One concrete route is to separate
mandatory error correction from discretionary stylistic refinement at
annotation time. Filtering after the fact does not work in one pass:
dropping stylistic edits wholesale leaves the rest just as inconsistent
(Appendix~\ref{sec:appendix_filter}). A promising future direction is
therefore an iterative loop that filters the post-edits, retrains, and
re-evaluates, repeating until the model becomes trainable or a retry
budget runs out; our diagnostic supplies the stopping rule such a loop
needs, since each round can simply check whether the $\lambda$ curve
has turned U-shaped. Deeper fixes such as MT
customization, task decomposition, or edit-type-conditioned generation
are future work. Our first attempt at the last, edit-level tagging at
training time, fails (Appendix~\ref{sec:appendix_failed}).

\section{Conclusion}
\label{sec:conclusion}
When APE fails to improve low-resource MT, the score alone cannot inform whether the model needs more training or the training data is too inconsistent to learn from. Our constrained-decoding diagnostic answers this without retraining or annotation: we sweep a single knob, an edit-distance penalty~$\lambda$ that drives the model from free editing (at $\lambda{=}0$) toward simply copying the MT output (at large $\lambda$), and read off whether the model learned net-positive edits and whether its confidence tracks edit quality. On English--Sinhala it exposes \textbf{Binary Collapse} and \textbf{Confident Miscalibration}, and both replicate on English--Marathi and English--Tamil. The evidence converges on our hypothesis that the heterogeneous post-edit distribution shapes both failure modes; model capacity, MT quality, and language family explain them less. Where genuine corrections survive training, a static constraint turns the same sweep into a free inference-time gain. Where they do not, the diagnostic reframes the failure as a data problem: test whether the post-edits form a learnable signal before investing in more data or compute. We release the first English--Sinhala and a new English--Tamil \hf{https://huggingface.co/datasets/isuruwijesiri/confident-but-wrong}{APE datasets} with all \gh{https://github.com/IsuruMaduranga/confident-but-wrong}{code}.

\raggedbottom
\section*{Limitations}

We identify six limitations. \emph{First, language coverage}: our findings are established on English--Sinhala and validated cross-lingually on English--Marathi (Indo-Aryan) and English--Tamil (Dravidian); additional language families (e.g., Turkish, Swahili) and the canonical high-resource APE setting (English--German) remain future work. \emph{Second, TER and the paraphrase concern}: the binary edit-distance penalty ($d{=}0$ if the token appears in the MT window, $1$ otherwise) penalizes a valid paraphrase as heavily as an error, and TER itself does not credit semantically equivalent rewrites. The token-level confidence finding (Table~\ref{tab:confidence}) is independent of TER, however: it directly measures the probability and entropy distributions of the model's edit decisions and shows that the model is no less confident on unnecessary edits than on correct ones ($p{\approx}\true{0.80}$ vs.\ $p{\approx}\true{0.81}$). So the central calibration result is robust to the paraphrase concern, even though TER itself is imperfect. \emph{Third, semantic and human evaluation}: we report TER, BLEU, and chrF++ throughout, and the diagnostic's curve-shape distinction is verified under all three (Appendix~\ref{sec:appendix_metric_robustness}), but human adequacy/fluency judgements and semantic-similarity QE on the predicted post-edits remain future work. \emph{Fourth, calibration measures}: our analysis covers token-level probability and entropy plus a sequence-level signal ($\lambda_{\text{seq-NLL}}$, Table~\ref{tab:constrained}), which agrees with the token-level ones; ensemble-based agreement measures remain untested and may provide better-calibrated trust signals. \emph{Fifth, dataset scale}: our datasets (${\sim}$\true{66k} English--Sinhala, ${\sim}$\true{39k} English--Tamil) are the largest for these language pairs but remain modest by WMT standards. \emph{Sixth, dataset-artifact framing}: our findings characterize APE under the \emph{inconsistent} post-edits of low-budget translation pipelines (Section~\ref{sec:why_collapse}); a reader might mistake this inconsistency for a removable data artifact rather than an intrinsic property of the APE task. But the conflicting corrections are often individually defensible (different post-editors make different yet legitimate discretionary choices), which is why naive filtering does not remove the inconsistency (Appendix~\ref{sec:appendix_filter}). Reducing it would require annotation protocols that separate mandatory error correction from discretionary refinement (Section~\ref{sec:action}), at some cost to ecological validity. The diagnostic remains useful under either framing, as a test of whether a given dataset supports learnable APE.

\section*{Ethics Statement}

The English--Sinhala and English--Tamil data underlying our APE experiments were produced as part of the EnSiTa dataset-creation effort~\citep{ensita2026}, which documents the translator recruitment, compensation, and quality-control protocol and confirms that the data contains no personally identifiable information; post-editors there were compensated at rates that meet or exceed standard local professional-translation wages. Our work adds only automated cleaning to derive APE triplets (Appendix~\ref{sec:appendix_preprocessing}) and involves no new human data collection. We note that APE systems, including ours, may propagate or amplify biases present in the underlying MT system; users should exercise caution when deploying APE for sensitive content. The human annotators who contributed to this study were compensated for their work.

\section*{Licenses and Use of Artifacts}

We use Gemma~3 1B and 4B~\citep{Gemma3} under the Gemma Terms of Use, NLLB-600M and NLLB-1.3B~\citep{nllbteam2022languageleftbehindscaling} under CC-BY-NC 4.0, and the WMT22 English--Marathi APE data~\citep{bhattacharyya-etal-2022-findings} under its shared-task license; all such use is for non-commercial research on automatic post-editing, consistent with each artifact's stated intended use. We release our English--Sinhala (${\sim}$\true{66k}) and English--Tamil (${\sim}$\true{39k}) APE datasets under CC-BY-SA 4.0 for research and academic use in machine translation, post-editing, and related NLP tasks; this is compatible with the conditions under which the underlying publicly available source texts and MT outputs were accessed. The D5 and English--Tamil NLLB-MT contrastive subsets contain NLLB-1.3B outputs, and downstream commercial redistribution of those subsets should additionally observe NLLB's CC-BY-NC 4.0 terms. Our code is available on \gh{https://github.com/IsuruMaduranga/confident-but-wrong}{GitHub}, and the released datasets are available on \hf{https://huggingface.co/datasets/isuruwijesiri/confident-but-wrong}{Hugging Face} (gated only to deter automated scraping, with access granted promptly to any researcher on request), under the terms above.

\section*{Acknowledgments}

This research was supported by a Google Diversity and Inclusion grant received by Surangika Ranathunga and Nisansa de Silva.

\section*{Use of AI Assistants}

We used a frontier general-purpose LLM assistant for iterative feedback on the paper (argument structure, clarity, and framing), for \LaTeX{} editing, and for developing experiment scripts; a second frontier LLM was used for code generation. GPT-5.2~\citep{openai2025gpt5} was used for the automated edit-type classification described in Section~\ref{sec:binary_collapse} and Appendix~\ref{sec:appendix_edit_classification}, and for the in-context-learning probes in Appendix~\ref{sec:appendix_prompts}. No AI system was used to generate experimental results, statistical analyses, or scientific claims, except where an AI system was itself the object of study. All AI-assisted text and code were reviewed, validated, and where necessary rewritten by the authors, who take full responsibility for the content.

\bibliography{custom}

\appendix

\section{In-Context Learning Results}
\label{sec:appendix_prompts}

\subsection{In-Context Learning Probe}
\label{sec:appendix_icl}

To test whether model capacity resolves the failure, we evaluate GPT-5.2~\citep{openai2025gpt5} on the same \true{200} stratified D1 samples used for edit-type classification (Section~\ref{sec:edit_types}) as an upper-bound probe under three prompting strategies: zero-shot, fixed few-shot (5 examples explicitly showing ``copy'' behavior), and retrieval-augmented (RAG, $k{=}5$ via TF-IDF). All strategies degrade substantially over the MT baseline (TER~\true{34.93}--\true{37.92} vs.\ MT~\true{28.82}; Table~\ref{tab:gpt5_results}). RAG performs \emph{worse} than both fixed few-shot and zero-shot, possibly because its retrieved examples reflect heterogeneous editing patterns. This probe points to task-level rather than capacity-level failure: even a frontier LLM with carefully chosen examples cannot distinguish necessary from unnecessary edits under this training signal.

\subsection{Detailed Results}

All experiments use temperature 0.0 and \texttt{max\_completion\_tokens}$=$512.

\begin{table}[!ht]
\centering
\small
\begin{tabular}{lccc}
\toprule
\textbf{Strategy} & \textbf{TER}$\downarrow$ & \textbf{BLEU}$\uparrow$ & \textbf{chrF++}$\uparrow$ \\
\midrule
MT Baseline & \true{28.82} & \true{63.24} & \true{77.96} \\
\midrule
Zero-shot & \true{37.51} & \true{54.90} & \true{72.84} \\
Fixed few-shot & \true{34.93} & \true{57.61} & \true{74.44} \\
RAG ($k{=}5$) & \true{37.92} & \true{54.47} & \true{72.06} \\
Two-stage (gate+edit) & \true{36.75} & \true{55.28} & \true{73.47} \\
\bottomrule
\end{tabular}
\caption{GPT-5.2 APE results on \true{200} stratified D1 test samples (same samples as edit-type analysis, Section~\ref{sec:edit_types}). All prompting strategies degrade substantially over the MT baseline. RAG performs \emph{worse} than both fixed few-shot and zero-shot; the two-stage decomposition (Section~\ref{sec:appendix_two_stage}) is intermediate and shows the same failure mode for structured prompting.}
\label{tab:gpt5_results}
\end{table}

Fixed few-shot performs best (TER~\true{34.93}) by explicitly showing ``copy'' behavior, but still degrades by \true{6.12} TER points over the MT baseline. Zero-shot degrades further (TER~\true{37.51}).

The most revealing result is that RAG performs \emph{worst} of all three strategies (TER~\true{37.92}), even below zero-shot. Retrieved examples faithfully reflect the training distribution, where 69.3\% of post-edits involve non-trivial changes (Section~\ref{sec:edit_types}), teaching the model the post-editor's heterogeneous editing patterns and amplifying over-editing. More context about the task appears to worsen performance, consistent with the training signal itself being inconsistent.

\paragraph{Base model zero-shot.} To confirm that fine-tuning is necessary (rather than prompt engineering being sufficient with a local model), we evaluate the base Gemma 3 1B-it and 4B-it with the same zero-shot APE prompt used for GPT-5.2 (Section~\ref{sec:appendix_icl}) on the full D1 test set ($n{=}\true{716}$). The 1B base model achieves TER~\true{190.31} and the 4B achieves TER~\true{93.81}, compared to the MT baseline of \true{24.98} (Table~\ref{tab:cross_model}): both generate fluent but largely unrelated Sinhala text, unable to perform APE without task-specific training. Fine-tuning reduces these to TER~\true{25.85} (1B) and \true{25.41} (4B), confirming that the models \emph{do} learn the task; the problem is not capacity but \emph{what} they learn (Binary Collapse).

\subsection{Two-Stage Decomposition (Gate + Edit)}
\label{sec:appendix_two_stage}

A structured-prompting strategy more sophisticated than zero-shot, few-shot, or RAG is to decompose APE into two independent decisions: \emph{(Stage~1)}~given (source, MT), decide whether the MT needs editing at all; \emph{(Stage~2)}~if Stage~1 says yes, generate a post-edit. We run both stages with GPT-5.2 at temperature~0 on the same \true{200} stratified D1 samples. Stage~1 is asked to output a single token \texttt{EDIT} or \texttt{KEEP}; Stage~2 uses the same prompt as the zero-shot APE setting (Section~\ref{sec:appendix_icl}). When Stage~1 outputs \texttt{KEEP}, we emit the MT unchanged.

\begin{table}[!ht]
\centering
\small
\begin{tabular}{lcc}
\toprule
\textbf{Metric} & \textbf{Value} & \textbf{Baseline} \\
\midrule
TER $\downarrow$ & \true{36.75} & MT \true{28.82} \\
\midrule
Gate precision & \true{0.75} & ``always edit'' \true{0.73} \\
Gate recall & \true{0.99} & -- \\
Gate F1 & \true{0.85} & -- \\
Edits called & \true{192/200} (\true{96\%}) & -- \\
\bottomrule
\end{tabular}
\caption{Two-stage decomposition with GPT-5.2 on \true{200} stratified D1 test samples. Confusion matrix: TP=\true{144}, FP=\true{48}, TN=\true{6}, FN=\true{2}. Gold edit rate (positive class proportion) is \true{0.73}; the ``always edit'' baseline therefore achieves gate precision \true{0.73} by definition. GPT-5.2's gate barely improves on this (\true{0.75}), and triggers an edit on \true{96\%} of inputs, so the pipeline collapses to vanilla zero-shot APE for almost all samples.}
\label{tab:two_stage}
\end{table}

The diagnostically important number is the gate precision (Table~\ref{tab:two_stage}): \true{0.75}, only \true{2} points above the trivial ``always edit'' baseline (gold edit rate \true{0.73}). GPT-5.2 says \texttt{EDIT} on \true{192/200} samples (recall \true{0.99}, precision \true{0.75}, F1 \true{0.85}), so the pipeline collapses to vanilla zero-shot APE on \true{96\%} of inputs. The resulting TER (\true{36.75}) is intermediate between zero-shot (\true{37.51}) and fixed few-shot (\true{34.93}), and still \true{7.93} TER points above the MT baseline. This gate failure is the same one XLM-R showed in Appendix~\ref{sec:appendix_qe} (AUROC~\true{0.745}, gating precision below \true{55\%} at the 90th-percentile threshold): a second, independent indication that, for the tested models and prompting strategies, the (source, MT) feature space does not provide a reliable edit-need signal.

\paragraph{Stage~1 system prompt.}
\begin{quote}
\small\ttfamily
You are a gating system for English-to-Sinhala automatic post-editing. You will receive an English source sentence and its Sinhala machine translation (MT). Decide whether the MT requires any post-editing.

\medskip
Output exactly one token:
\begin{itemize}
\item \texttt{KEEP} if the MT is acceptable as-is (no genuine errors).
\item \texttt{EDIT} if the MT contains a genuine error (wrong meaning, mistranslation, missing content, ungrammatical form).
\end{itemize}

\medskip
Do not output anything else.
\end{quote}

\subsection{Few-Shot Examples and RAG Retrieval}

The fixed few-shot mode prepends 5 static examples (Table~\ref{tab:fewshot_examples}) as alternating user/assistant turns: 2 ``copy'' cases (TER$=$0), 2 minor edits, and 1 major edit. For RAG, $k{=}5$ training examples are retrieved per test sample using TF-IDF cosine similarity over English source sentences, formatted identically but with dynamically selected examples (TF-IDF fit on full D1 training set, $n{=}$\true{5,728}).

\begin{table*}[!t]
\centering
\footnotesize
\setlength{\tabcolsep}{4pt}
\begin{tabular}{cp{5.0cm}p{5.0cm}c}
\toprule
\textbf{\#} & \textbf{User (Source + MT)} & \textbf{Assistant (Post-Edit)} & \textbf{Type} \\
\midrule
1 & Source: When he saw who his visitor was his eyes brightened. \newline MT: තම අමුත්තා කවුදැයි දුටු විට ඔහුගේ දෑස් දීප්තිමත් විය. & තම අමුත්තා කවුදැයි දුටු විට ඔහුගේ දෑස් දීප්තිමත් විය.
 & Copy \\
\midrule
2 & Source: They went to the football field and started their game. \newline MT: ඔවුන් පාපන්දු පිටියට ගොස් ඔවුන්ගේ ක්‍රීඩාව ආරම්භ කළහ. & ඔවුන් පාපන්දු පිටියට ගොස් ඔවුන්ගේ ක්‍රීඩාව ආරම්භ කළහ. & Copy \\
\midrule
3 & Source: Monte Cristo bowed, and went to Madame de Villefort. \newline MT: මොන්ටේ ක්‍රිස්ටෝ හිස නමා මැඩම් ඩි විල්ෆෝර්ට් වෙත ගියේය. & මොන්ටේ ක්‍රිස්ටෝ හිස නමා{\color{editcol},} මැඩම් ඩි විල්ෆෝර්ට් වෙත ගියේය. & Minor \\
\midrule
4 & Source: ``I do not know her name; but it is she, sir, it is she!'' \newline MT: ``මම ඇගේ නම දන්නේ නැහැ; නමුත් ඒ ඇයයි, සර්, මේ ඇයයි!'' & ``මම ඇගේ නම දන්නේ නැහැ; නමුත් ඒ ඇයයි, {\color{editcol}මහත්මයා, ඒ} මේ ඇයයි!'' & Minor \\
\midrule
5 & Source: He melted in the hot sun. \newline MT: ඔහු උණුසුම් හිරු තුළ දිය වී ගියේය. & ඔහු උණුසුම් හිරු {\color{editcol}එළියේ} දිය වී ගියේය. & Major \\
\bottomrule
\end{tabular}
\caption{Fixed few-shot examples for GPT-5.2. Examples 1--2 show ``copy'' behavior (TER$=$0), 3--4 show minor edits, and 5 shows a major edit. All from D1 training set.}
\label{tab:fewshot_examples}
\end{table*}

\subsection{System Prompts}

\paragraph{Zero-shot and fixed few-shot.} Both modes use the same system prompt:

\begin{quote}
\small\ttfamily
You are an Automatic Post-Editing system for English-to-Sinhala machine translation. You will receive an English source sentence and its Sinhala machine translation (MT). Your task:

1. If the MT is correct, output it EXACTLY as-is. Do not change anything.

2. If the MT has a genuine translation error (wrong meaning, mistranslation, missing content), fix ONLY that error.

3. Do NOT change style, register, word choice, loanwords, or word order when the meaning is already correct. These are stylistic preferences, not errors.

4. Make the MINIMUM number of changes needed. Fewer changes is better.

Output ONLY the post-edited Sinhala text, nothing else.
\end{quote}

\paragraph{RAG mode.} The RAG system prompt instructs the model to follow the post-editor's style from retrieved examples:

\begin{quote}
\small\ttfamily
You are an Automatic Post-Editing system for English-to-Sinhala machine translation. You will receive an English source sentence and its Sinhala machine translation (MT), along with examples of how a professional post-editor corrected similar sentences.

Follow the post-editor's style shown in the examples:

- If the examples show the post-editor kept the MT unchanged, you should also keep the MT unchanged for similar cases.

- If the examples show small targeted fixes, make only similar small fixes.

- Match the level of editing shown in the examples.

Output ONLY the post-edited Sinhala text, nothing else.
\end{quote}

\subsection{User Message Format}

For all modes, each test sample is formatted as:

\begin{quote}
\small\ttfamily
Source: \{source\}\\
MT: \{mt\}
\end{quote}

\section{Dataset Creation Details}
\label{sec:appendix_dataset_creation}
All English--Sinhala and English--Tamil data come from the EnSiTa corpus~\citep{ensita2026}.

\subsection{D1--D4 (Google Translate)}
The EnSiTa dataset-creation effort produced the D1--D4 source sentences (D1: literary texts, D2: news, D3: Wikipedia, D4: mathematics/health textbooks), Google Translate MT outputs, and professional Sinhala post-edits~\citep{ensita2026}. It documents source selection, translator recruitment, compensation, and quality control. We obtain the resulting triplets and construct APE splits through the cleaning pipeline below.

\subsection{D5 (From NLLB Corpus, NLLB MT)}
D5 is a controlled contrastive condition we construct for this study. English sentences from the NLLB parallel corpus~\citep{nllbteam2022languageleftbehindscaling} were cleaned and translated into Sinhala with NLLB-1.3B (distilled) to obtain a deliberately lower-quality MT, then post-edited under the same protocol as D1--D4.

All English--Sinhala datasets follow an 80/10/10 train/validation/test split.

\subsection{Preprocessing: From Released Dataset to Experimental Splits}
\label{sec:appendix_preprocessing}

The EnSiTa creation process yields, for each sentence, the source, an initial MT, and the professional post-edit; we extract these as (source, MT, PE) triplets for D1--D4 and the English--Tamil domains. We discard any item whose source text was itself revised during EnSiTa's editing: some items passed through several rounds of back-and-forth revision that also changed the source, violating the fixed-source assumption of APE (the source must be identical for the MT and the post-edit). The retained triplets, together with the D5 triplets we construct separately (Table~\ref{tab:ensi_dataset}), are cleaned into the experimental splits D1--D5 (Table~\ref{tab:dataset}) by the following pipeline, applied identically to all domains:

\begin{enumerate}
    \item \textbf{Remove empty/null rows}: drop any triplet where SRC, MT, or PE is missing or empty.
    \item \textbf{Deduplication}: remove exact (SRC, MT, PE) duplicates, keeping the first occurrence.
    \item \textbf{Length filtering}: remove triplets where any field has fewer than 5 or more than 2,000 characters.
    \item \textbf{PE/MT length ratio filtering}: remove triplets where $\text{len(PE)}/\text{len(MT)} \notin [0.5, 1.5]$, excluding extreme rewrites or truncations that would distort TER-based evaluation.
    \item \textbf{Shuffle and split}: shuffle with a fixed seed (42), then split 80/10/10 into train/validation/test.
\end{enumerate}

\noindent This pipeline removes fewer than 1\% of triplets for most domains. The released datasets include all triplets before cleaning, enabling researchers to apply their own preprocessing.

\begin{table}[t]
\centering
\small
\begin{tabular}{lrrl}
\toprule
\textbf{Domain} & \textbf{Triplets} & \textbf{Mean TER} & \textbf{Edit Dist.} \\
\midrule
Literature & \true{7,209} & \true{25.7} & \true{30/21/49} \\
News & \true{16,632} & \true{9.9} & \true{51/33/16} \\
Wikipedia & \true{21,839} & \true{14.9} & \true{25/45/30} \\
Maths & \true{6,487} & \true{23.3} & \true{29/26/45} \\
Health & \true{1,000} & \true{42.1} & \true{13/18/69} \\
NLLB Corpus & \true{12,915} & \true{61.0} & \true{2/5/93} \\
\midrule
\textbf{Total} & \true{66,082} & -- & -- \\
\bottomrule
\end{tabular}
\caption{English--Sinhala APE dataset release statistics. Mean TER is the corpus-level mean TER(MT, PE) over all triplets before cleaning. Edit Dist.\ shows none/minor/major percentages (TER thresholds: 0, 20). Literature, News, Wikipedia, Maths, and Health use Google Translate MT; NLLB Corpus uses NLLB-1.3B MT. Experimental splits (D1--D5) derived from these domains after cleaning are in Table~\ref{tab:dataset}.}
\label{tab:ensi_dataset}
\end{table}

\subsection{English--Tamil APE Dataset}
\label{sec:appendix_enta}

Alongside the English--Sinhala data, we derive an English--Tamil APE dataset spanning five domains (Table~\ref{tab:enta_dataset}) from the English--Tamil part of the same EnSiTa effort~\citep{ensita2026}: ${\sim}$39k (source, Google Translate MT, post-edit) triplets built with the same cleaning pipeline as D1--D4. Constrained decoding experiments use the News subset (\true{10,000} triplets); per-domain vanilla fine-tuning results are in Table~\ref{tab:enta_domain}. The remaining domains are released to support future APE research on Dravidian languages. Tamil, a Dravidian language, is typologically distinct from the Indo-Aryan languages in our main experiments (Sinhala, Marathi).

\begin{table}[t]
\centering
\small
\begin{tabular}{lrrl}
\toprule
\textbf{Domain} & \textbf{Triplets} & \textbf{Mean TER} & \textbf{Edit Dist.} \\
\midrule
Wiki & \true{19,864} & \true{12.7} & \true{41/37/23} \\
News & \true{10,000} & \true{20.2} & \true{44/20/36} \\
Literature & \true{3,501} & \true{26.8} & \true{42/13/45} \\
Maths & \true{3,423} & \true{23.6} & \true{25/30/44} \\
Health & \true{1,870} & \true{7.3} & \true{69/17/15} \\
\midrule
\textbf{Total} & \true{38,658} & -- & -- \\
\bottomrule
\end{tabular}
\caption{English--Tamil APE dataset statistics. Edit Dist.\ shows the percentage of none/minor/major edits (TER thresholds: 0, 20). All domains use Google Translate MT. Experiments in this paper use only News; remaining domains are released for future research.}
\label{tab:enta_dataset}
\end{table}

\section{Full Results Tables}
\label{sec:appendix_results}

\subsection{Per-Seed Breakdown}

Table~\ref{tab:full_seeds} gives the per-seed TER, BLEU, and chrF++ behind the three-seed means reported in the main text.

\begin{table*}[!t]
\centering
\footnotesize
\renewcommand{\arraystretch}{0.92}
\setlength{\tabcolsep}{4pt}
\begin{tabular}{ll|ccc|ccc|ccc}
\toprule
& & \multicolumn{3}{c|}{\textbf{Seed 42}} & \multicolumn{3}{c|}{\textbf{Seed 52}} & \multicolumn{3}{c}{\textbf{Seed 62}} \\
\textbf{Model} & \textbf{Method} & TER & BLEU & chrF++ & TER & BLEU & chrF++ & TER & BLEU & chrF++ \\
\midrule
\multirow{5}{*}{\rotatebox{90}{Gemma 1B}} & Vanilla & \true{25.84} & \true{67.34} & \true{79.58} & \true{25.84} & \true{67.06} & \true{79.65} & \true{25.88} & \true{67.07} & \true{79.49} \\
& HSA & \true{25.26} & \true{67.42} & \true{79.78} & \true{25.89} & \true{67.20} & \true{79.58} & \true{25.60} & \true{67.24} & \true{79.62} \\
& $\lambda_{\text{entropy}}$ & \true{25.81} & \true{67.30} & \true{79.55} & \true{25.65} & \true{67.24} & \true{79.71} & \true{25.69} & \true{67.14} & \true{79.54} \\
& $\lambda_{\text{pmt}}$ & \true{25.65} & \true{67.34} & \true{79.65} & \true{25.18} & \true{67.59} & \true{79.88} & \true{25.29} & \true{67.43} & \true{79.72} \\
& $\lambda_{\text{static}}$ & \true{24.97} & \true{67.31} & \true{79.86} & \true{25.07} & \true{67.43} & \true{79.86} & \true{24.91} & \true{67.39} & \true{79.86} \\
\midrule
\multirow{4}{*}{\rotatebox{90}{Gemma 4B}} & Vanilla & \true{25.15} & \true{67.43} & \true{79.88} & \true{25.36} & \true{67.46} & \true{79.84} & \true{25.72}$^\dagger$ & \true{67.21}$^\dagger$ & \true{79.61}$^\dagger$ \\
& $\lambda_{\text{entropy}}$ & \true{25.05} & \true{67.53} & \true{79.95} & -- & -- & -- & -- & -- & -- \\
& $\lambda_{\text{pmt}}$ & \true{24.90} & \true{67.76} & \true{79.97} & -- & -- & -- & -- & -- & -- \\
& $\lambda_{\text{static}}$ & \true{24.83} & \true{67.52} & \true{79.95} & -- & -- & -- & -- & -- & -- \\
\midrule
\multirow{4}{*}{\rotatebox{90}{NLLB}} & Vanilla & \true{31.10} & \true{61.77} & \true{77.30} & \true{30.44} & \true{62.25} & \true{77.54} & \true{30.80} & \true{61.92} & \true{77.29} \\
& $\lambda_{\text{entropy}}$ & \true{31.28} & \true{61.57} & \true{77.09} & -- & -- & -- & -- & -- & -- \\
& $\lambda_{\text{pmt}}$ & \true{29.97} & \true{62.58} & \true{77.68} & -- & -- & -- & -- & -- & -- \\
& $\lambda_{\text{static}}$ & \true{29.59} & \true{62.63} & \true{77.61} & -- & -- & -- & -- & -- & -- \\
\bottomrule
\end{tabular}
\caption{Full per-seed results on D1 (Literature). Most constrained decoding methods improve over vanilla fine-tuning, with static constraint consistently achieving the best results; entropy is the exception, trailing vanilla on NLLB. $^\dagger$Gemma 4B seed 62 diverged during QLoRA training; seed 72 used instead.}
\label{tab:full_seeds}
\end{table*}

\subsection{Token-Level Confidence Distribution}
\label{sec:appendix_confidence_dist}

\begin{figure*}[t!]
  \centering
  \includegraphics[width=0.93\textwidth]{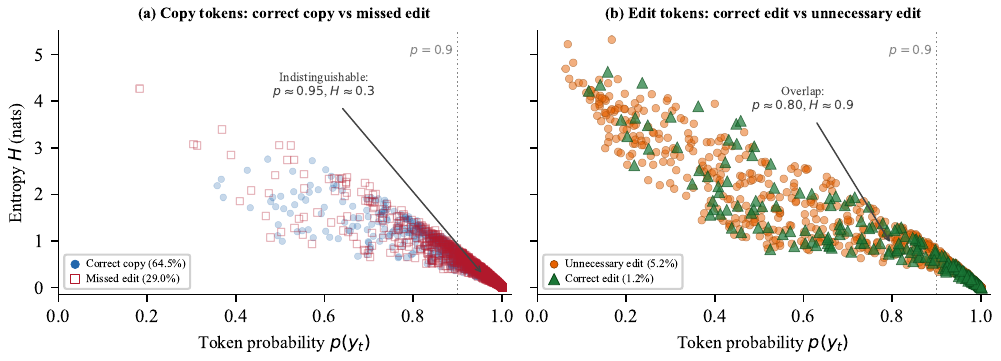}
  \caption{Token-level confidence vs.\ edit correctness for Gemma 3 1B on D1. \true{59.4\%} of unnecessary edits have $p > 0.9$, indistinguishable from the \true{54.8\%} of correct edits at $p > 0.9$: per-token probability does not separate net-positive edits from unnecessary ones. The \true{84.8\%} of missed edits at $p > 0.9$ shows confidence is equally uninformative about whether to edit at all.}
  \label{fig:confidence_scatter}
\end{figure*}

Figure~\ref{fig:confidence_scatter} shows the full per-token probability--entropy distribution summarized in Table~\ref{tab:confidence}.

\subsection{Temperature Scaling Results}
\label{sec:appendix_temperature}

Table~\ref{tab:temperature} reports the effect of temperature scaling, which worsens unconstrained generation while preserving the constraint ordering. Recalibrating confidence therefore does not fix the failure. Only the static constraint, which ignores model confidence, stays near the MT baseline.

\begin{table}[!ht]
\centering
\small
\begin{tabular}{lcc}
\toprule
\textbf{Method} & \textbf{TER ($T{=}1$)} & \textbf{TER ($T{=}2$)} \\
\midrule
MT Baseline & \true{24.98} & \true{24.98} \\
Vanilla FT ($\lambda{=}0$) & \true{25.84} & \true{27.63} \\
\midrule
$\lambda_{\text{entropy}}$ (best) & \true{25.81} & \true{27.40} \\
$\lambda_{\text{pmt}}$ (best) & \true{25.65} & \true{25.66} \\
\rowcolor{green!10}
$\lambda_{\text{static}}$ (best) & \textbf{\true{24.97}} & \textbf{\true{25.08}} \\
\bottomrule
\end{tabular}
\caption{Effect of temperature scaling ($T^*{=}\true{2.0}$, tuned on val NLL) on D1, seed~42. Calibration \emph{worsens} unconstrained generation (\true{25.84}${\to}$\true{27.63}) and prevents the static constraint from recovering the MT baseline. The ordering Static~${>}$~PMT~${>}$~Entropy is preserved.}
\label{tab:temperature}
\end{table}

\subsection{Lambda Sensitivity (Full Results)}

Tables~\ref{tab:lambda_sweep},~\ref{tab:lambda_sweep_d5}, and~\ref{tab:nllb_d5_lambda} report the full $\lambda$ sweeps for D1 and D5 (Gemma and NLLB-600M); Tables~\ref{tab:enmr_lambda} and~\ref{tab:enta_lambda} cover English--Marathi and English--Tamil News; Tables~\ref{tab:lambda_sweep_d2}--\ref{tab:lambda_sweep_d4} cover D2--D4.

\begin{table}[!ht]
\centering
\small
\begin{tabular}{lccc}
\toprule
\textbf{$\lambda$} & \textbf{Static} & \textbf{Entropy} & \textbf{PMT} \\
\midrule
0.0 & \true{25.84} & -- & -- \\
0.1 & \true{25.81} & -- & -- \\
0.3 & \true{25.66} & -- & -- \\
0.5 & \true{25.65} & \true{25.81} & \true{25.79} \\
1.0 & \true{25.00} & \true{25.81} & \true{25.69} \\
2.0 & \true{24.98} & \true{25.81} & \true{25.65} \\
2.5 & \textbf{\true{24.97}} & -- & -- \\
3.0 & \true{24.96} & \true{25.81} & \true{25.76} \\
\bottomrule
\end{tabular}
\caption{TER vs.\ $\lambda$ sweep (D1, Gemma 3 1B, seed 42). Static converges to MT-level quality at $\lambda{\geq}1.0$. Entropy is flat (${\sim}$25.81 at all $\lambda$): the constraint barely activates. PMT shows a real curve with minimum at $\lambda{=}2.0$, but never approaches the MT baseline.}
\label{tab:lambda_sweep}
\end{table}

\begin{table}[!ht]
\centering
\small
\begin{tabular}{lcc}
\toprule
\textbf{$\lambda$} & \textbf{TER} $\downarrow$ & \textbf{$\Delta$ vs MT} \\
\midrule
MT Baseline & \true{59.19} & -- \\
\midrule
0.0 & \true{57.16} & \true{+2.03} \\
0.1 & \true{56.99} & \true{+2.20} \\
0.3 & \true{57.12} & \true{+2.07} \\
0.5 & \true{57.03} & \true{+2.16} \\
0.7 & \true{56.75} & \true{+2.44} \\
1.0 & \true{56.11} & \true{+3.08} \\
1.5 & \true{55.98} & \true{+3.21} \\
\rowcolor{green!10}
\textbf{2.0} & \textbf{\true{55.21}} & \textbf{\true{+3.98}} \\
2.5 & \true{56.05} & \true{+3.14} \\
3.0 & \true{55.47} & \true{+3.72} \\
5.0 & \true{58.55} & \true{+0.64} \\
\bottomrule
\end{tabular}
\caption{TER vs.\ $\lambda$ sweep (D5, Gemma 3 1B, seed 42, static constraint). Unlike D1's monotonic curve, D5 shows a U-shape: TER dips below both vanilla FT (\true{56.89}) and MT baseline at $\lambda{=}2.0$, then rises at $\lambda{=}5.0$ as the constraint forces copying of low-quality MT.}
\label{tab:lambda_sweep_d5}
\end{table}

\begin{table}[!ht]
\centering
\small
\begin{tabular}{lcc}
\toprule
\textbf{$\lambda$} & \textbf{TER} $\downarrow$ & \textbf{$\Delta$ vs MT} \\
\midrule
MT Baseline & \true{59.19} & -- \\
\midrule
$-$5.0 & \true{77.37} & \true{$-$18.18} \\
$-$3.0 & \true{71.24} & \true{$-$12.05} \\
$-$2.0 & \true{66.05} & \true{$-$6.87} \\
$-$1.0 & \true{61.10} & \true{$-$1.91} \\
$-$0.5 & \true{58.87} & \true{+0.31} \\
\rowcolor{green!10}
\textbf{0.0} & \textbf{\true{57.14}} & \textbf{\true{+2.05}} \\
1.0 & \true{57.34} & \true{+1.85} \\
1.5 & \true{57.34} & \true{+1.85} \\
2.0 & \true{57.71} & \true{+1.47} \\
3.0 & \true{59.94} & \true{$-$0.76} \\
6.0 & \true{64.59} & \true{$-$5.41} \\
14.0 & \true{71.34} & \true{$-$12.15} \\
\bottomrule
\end{tabular}
\caption{TER vs.\ $\lambda$ sweep (D5, NLLB-600M, seed 42, static constraint). The curve is U-shaped with its minimum at $\lambda{=}0$ (vanilla FT). Negative $\lambda$ penalizes MT tokens (encouraging hallucination); positive $\lambda$ forces MT copying. Both directions degrade performance, with the negative direction steeper, consistent with low-quality MT structure providing useful scaffolding. Values saturate at $|\lambda|{\geq}12$.}
\label{tab:nllb_d5_lambda}
\end{table}

\begin{table}[!ht]
\centering
\small
\begin{tabular}{lccc}
\toprule
\textbf{$\lambda$} & \textbf{$\lambda_{\text{static}}$} & \textbf{$\lambda_{\text{pmt}}$} & \textbf{$\lambda_{\text{entropy}}$} \\
\midrule
MT Baseline & \multicolumn{3}{c}{\true{22.93}} \\
Vanilla FT & \multicolumn{3}{c}{\true{22.24}} \\
\midrule
0.5 & \true{21.98} & \true{22.25} & \true{22.21} \\
1.0 & \true{21.83} & \true{22.19} & \true{22.15} \\
\rowcolor{green!10}
2.0 & \textbf{\true{21.73}} & \true{22.05} & \true{22.10} \\
3.0 & \true{21.83} & \true{21.90} & \true{22.09} \\
5.0 & \true{22.46} & \true{21.83} & \true{22.06} \\
\bottomrule
\end{tabular}
\caption{TER vs.\ $\lambda$ sweep (English--Marathi, Gemma 3 1B, seed 42). All three constraint methods shown. The static constraint produces a U-shaped curve with best TER~\true{21.73} at $\lambda{=}2.0$. The ordering Static~$>$~PMT~$>$~Entropy holds at every $\lambda$ value tested. Entropy TER is nearly flat across $\lambda$ (same pattern as D1 and D5), confirming that the entropy-scaled constraint barely activates.}
\label{tab:enmr_lambda}
\end{table}

\begin{table}[!ht]
\centering
\small
\begin{tabular}{lc}
\toprule
\textbf{$\lambda$} & \textbf{$\lambda_{\text{static}}$ TER} \\
\midrule
MT Baseline & \true{20.57} \\
Vanilla FT & \true{21.77} \\
\midrule
0.5 & \true{24.24} \\
1.0 & \true{23.34} \\
2.0 & \true{21.46} \\
3.0 & \true{20.89} \\
5.0 & \true{20.68} \\
\bottomrule
\end{tabular}
\caption{TER vs.\ $\lambda$ sweep (English--Tamil News, Gemma 3 1B, seed 42, static constraint). The curve is monotonic, approaching but never crossing the MT baseline (\true{20.57}). Plotted in Figure~\ref{fig:enta_contrastive}(a).}
\label{tab:enta_lambda}
\end{table}

\begin{table}[!ht]
\centering
\small
\begin{tabular}{lcc}
\toprule
\textbf{$\lambda$} & \textbf{TER} $\downarrow$ & \textbf{$\Delta$ vs MT} \\
\midrule
MT Baseline & \true{47.54} & -- \\
Vanilla FT & \true{50.90} & \true{$-$3.36} \\
\midrule
0.5 & \true{53.42} & \true{$-$5.87} \\
1.0 & \true{51.08} & \true{$-$3.54} \\
2.0 & \true{48.16} & \true{$-$0.62} \\
\rowcolor{green!10}
\textbf{3.0} & \textbf{\true{47.72}} & \textbf{\true{$-$0.18}} \\
4.0 & \true{48.11} & \true{$-$0.57} \\
5.0 & \true{48.48} & \true{$-$0.94} \\
7.0 & \true{48.36} & \true{$-$0.82} \\
10.0 & \true{48.20} & \true{$-$0.66} \\
\bottomrule
\end{tabular}
\caption{TER vs.\ $\lambda$ sweep (English--Tamil News with NLLB-1.3B MT, Gemma 3 1B, seed 42, static constraint). Same post-edits as Table~\ref{tab:enta_lambda} but with synthetic NLLB MT. The curve is U-shaped with minimum at $\lambda{=}3.0$, then plateaus at ${\sim}$48.2 for $\lambda{\geq}7$. Plotted in Figure~\ref{fig:enta_contrastive}(b).}
\label{tab:enta_synth_lambda}
\end{table}

\begin{figure*}[!ht]
  \centering
  \includegraphics[width=0.93\textwidth]{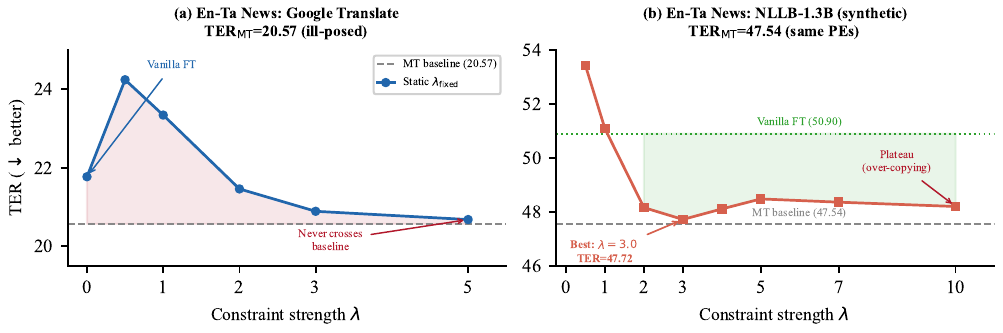}
  \caption{English--Tamil News $\lambda$ curves: same post-edits, different MT. \textbf{(a)}~Google Translate MT (TER$_{\mathrm{MT}}{=}\true{20.57}$): monotonic, never crosses MT baseline. \textbf{(b)}~NLLB-1.3B MT (TER$_{\mathrm{MT}}{=}\true{47.54}$): U-shaped with minimum at $\lambda{=}\true{3.0}$ (TER$\,{=}\,\true{47.72}$). The within-language contrast is consistent with a task-difficulty effect rather than a language-specific effect. Corresponding sweep values in Tables~\ref{tab:enta_lambda} and~\ref{tab:enta_synth_lambda}.}
  \label{fig:enta_contrastive}
\end{figure*}

\begin{table}[!ht]
\centering
\small
\setlength{\tabcolsep}{5pt}
\begin{tabular}{lcccc}
\toprule
\textbf{Domain} & \textbf{MT TER} & \textbf{FT TER} & \textbf{$\Delta$TER} & \textbf{Triplets} \\
\midrule
Wiki & \true{12.67} & \true{12.95} & \true{$-$0.28} & \true{23,880} \\
News & \true{20.57} & \true{21.77} & \true{$-$1.21} & \true{10,000} \\
Combined$^\dagger$ & \true{19.07} & \true{18.96} & \true{$+$0.11} & \true{8,794} \\
\bottomrule
\end{tabular}
\caption{English--Tamil per-domain vanilla fine-tuning results (Gemma 3 1B, seed 42). MT source is Google Translate for all domains. Constrained decoding experiments are conducted only on the News domain (Section~\ref{sec:validation}). $^\dagger$Literature (3,501), Maths (3,423), and Health (1,870) are combined into a single split because individually each is too small for stable fine-tuning; Literature alone failed to converge.}
\label{tab:enta_domain}
\end{table}

\begin{table}[!ht]
\centering
\small
\begin{tabular}{lcc}
\toprule
\textbf{$\lambda$} & \textbf{TER} $\downarrow$ & \textbf{$\Delta$ vs MT} \\
\midrule
MT Baseline & \true{7.93} & -- \\
\midrule
0.0 & \true{7.91} & \true{+0.02} \\
0.5 & \true{9.17} & \true{$-$1.24} \\
1.0 & \true{8.32} & \true{$-$0.39} \\
2.0 & \true{8.25} & \true{$-$0.33} \\
2.5 & \true{8.12} & \true{$-$0.19} \\
3.0 & \true{8.12} & \true{$-$0.20} \\
\bottomrule
\end{tabular}
\caption{TER vs.\ $\lambda$ sweep (D2 News, Gemma 3 1B, seed 42, static constraint). Vanilla FT already marginally improves over MT; the constraint monotonically worsens TER, confirming the model already makes net-positive edits and no inference-time correction is needed.}
\label{tab:lambda_sweep_d2}
\end{table}

\begin{table}[!ht]
\centering
\small
\begin{tabular}{lcc}
\toprule
\textbf{$\lambda$} & \textbf{TER} $\downarrow$ & \textbf{$\Delta$ vs MT} \\
\midrule
MT Baseline & \true{14.89} & -- \\
\midrule
0.0 & \true{14.88} & \true{+0.01} \\
0.5 & \true{15.01} & \true{$-$0.12} \\
1.0 & \true{14.71} & \true{+0.19} \\
2.0 & \true{14.45} & \true{+0.45} \\
2.5 & \true{14.45} & \true{+0.44} \\
3.0 & \true{14.52} & \true{+0.38} \\
\bottomrule
\end{tabular}
\caption{TER vs.\ $\lambda$ sweep (D3 Wikipedia, Gemma 3 1B, seed 42, static constraint). The constraint beats the MT baseline at $\lambda{\geq}1.0$, with best TER~\true{14.45} at $\lambda{=}2.0$ ($\Delta$TER${=}\true{+0.45}$). Mild U-shape with uptick at $\lambda{=}3.0$.}
\label{tab:lambda_sweep_d3}
\end{table}

\begin{table}[!ht]
\centering
\small
\begin{tabular}{lcc}
\toprule
\textbf{$\lambda$} & \textbf{TER} $\downarrow$ & \textbf{$\Delta$ vs MT} \\
\midrule
MT Baseline & \true{24.30} & -- \\
\midrule
0.0 & \true{24.99} & \true{$-$0.68} \\
0.5 & \true{25.93} & \true{$-$1.63} \\
1.0 & \true{24.57} & \true{$-$0.27} \\
2.0 & \true{24.09} & \true{+0.21} \\
2.5 & \true{24.58} & \true{$-$0.27} \\
3.0 & \true{24.64} & \true{$-$0.34} \\
\bottomrule
\end{tabular}
\caption{TER vs.\ $\lambda$ sweep (D4 Maths/Health, Gemma 3 1B, seed 42, static constraint). Despite vanilla FT degrading TER by $\true{0.68}$, the U-shaped curve reveals genuine correctable errors, with best TER~\true{24.09} at $\lambda{=}2.0$ ($\Delta$TER${=}\true{+0.21}$, beats MT baseline).}
\label{tab:lambda_sweep_d4}
\end{table}

\subsection{Domain Robustness}
\label{sec:appendix_domain}

\begin{table}[!ht]
\centering
\small
\resizebox{\columnwidth}{!}{%
\begin{tabular}{llccc}
\toprule
\textbf{Dataset} & \textbf{System} & \textbf{TER} $\downarrow$ & \textbf{BLEU} $\uparrow$ & \textbf{$\Delta$TER} \\
\midrule
\multirow{3}{*}{D2 (News)} & MT Baseline & \true{7.93} & \true{89.25} & -- \\
& Vanilla FT & \true{7.91} & \true{89.56} & \true{+0.02} \\
& $\lambda_{\text{static}}$ & \true{8.12} & \true{89.39} & \true{$-$0.19} \\
\midrule
\multirow{3}{*}{D3 (Wiki)} & MT Baseline & \true{14.89} & \true{79.11} & -- \\
& Vanilla FT & \true{14.88} & \true{78.95} & \true{+0.01} \\
& $\lambda_{\text{static}}$ & \true{14.45} & \true{79.20} & \true{+0.45} \\
\midrule
\multirow{3}{*}{D4 (Math/H)} & MT Baseline & \true{24.30} & \true{71.18} & -- \\
& Vanilla FT & \true{24.99} & \true{69.28} & \true{$-$0.68} \\
& $\lambda_{\text{static}}$ & \true{24.09} & \true{70.43} & \true{+0.21} \\
\midrule
\multirow{3}{*}{D1 (Lit.)} & MT Baseline & \true{24.98} & \true{67.38} & -- \\
& Vanilla FT & \true{25.85} & \true{67.16} & \true{$-$0.87} \\
& $\lambda_{\text{static}}$ & \true{24.98} & \true{67.38} & \true{0.00} \\
\bottomrule
\end{tabular}}
\caption{Domain robustness (Gemma 3 1B, seed 42). Vanilla FT degrades TER on D1 ($-$0.87) and D4 ($-$0.68), but the static constraint reveals different underlying patterns: D4 shows a U-shaped curve that beats MT ($\Delta$TER${=}\true{+0.21}$), while D1 only recovers to MT level. D3's constraint similarly reveals correctable errors ($\Delta$TER${=}\true{+0.45}$). D2 is the opposite case: vanilla FT already marginally improves, and any constraint worsens TER ($-$0.19), so no tested constraint improves this result. Only on D1 (Literature) do no net-positive edits survive.}
\label{tab:domain}
\end{table}

\begin{figure*}[t]
  \centering
  \includegraphics[width=\textwidth]{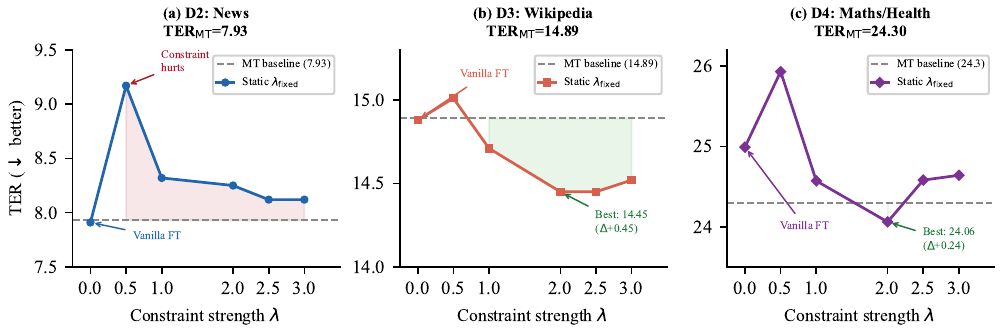}
  \caption{TER vs.\ $\lambda$ curves for D2--D4 (Gemma 3 1B, seed 42, static constraint). \textbf{Left:} D2 (News, TER$_{\mathrm{MT}}{=}\true{7.93}$): vanilla FT marginally improves over MT; the constraint monotonically worsens TER, confirming the model's edits are already net-positive. \textbf{Center:} D3 (Wikipedia, TER$_{\mathrm{MT}}{=}\true{14.89}$): vanilla FT shows negligible change ($\Delta$TER${=}\true{+0.01}$), but the static constraint reveals genuine correctable errors, reaching TER${\,=\,}\true{14.45}$ at $\lambda{=}\true{2.0}$ ($\Delta$TER${=}\true{+0.45}$); mild U-shape with uptick at $\lambda{=}3.0$. \textbf{Right:} D4 (Maths/Health, TER$_{\mathrm{MT}}{=}\true{24.30}$): vanilla FT degrades TER ($-$0.68), but unlike D1 the curve is U-shaped with best TER${\,=\,}\true{24.09}$ at $\lambda{=}\true{2.0}$ ($\Delta$TER${=}\true{+0.21}$), revealing genuine correctable errors despite similar MT quality.}
  \label{fig:lambda_d2_d3_d4}
\end{figure*}

Vanilla fine-tuning degrades TER on D1 (Literature, $\true{-0.87}$) and D4 (Maths/Health, $\true{-0.68}$), the two domains with higher MT TER (${\sim}$\true{24}--\true{25}). On D2 (News, $\true{+0.02}$) and D3 (Wiki, $\true{+0.01}$), where MT quality is already high (TER~\true{8} and \true{15}), it barely changes TER. The $\lambda$ curves, however, reveal distinctions that these aggregate scores hide (Figure~\ref{fig:lambda_d2_d3_d4}): D4's U-shaped curve beats the MT baseline at $\lambda{=}2.0$ ($\Delta$TER${=}\true{+0.21}$), whereas D1's monotonic curve only recovers to the MT level. D3 likewise dips below the MT baseline ($\Delta$TER${=}\true{+0.45}$, Table~\ref{tab:lambda_sweep_d3}). D2 behaves differently: vanilla fine-tuning already improves marginally over MT, and any constraint worsens TER (Table~\ref{tab:lambda_sweep_d2}), so the minimum sits at $\lambda{=}0$. This is consistent with net-positive editing where the edits are mostly mechanical (punctuation, spelling). Together, these curves support the hypothesis that Literature's style-transfer post-edits contribute to the failure.

\subsection{Seed Stability}

Signal-agnostic constrained methods (static, entropy) are \true{3--4$\times$} more stable than HSA ($\sigma \leq \true{0.09}$ vs.\ $\true{0.32}$; Table~\ref{tab:full_seeds}), as expected for inference-time interventions that add no training stochasticity. PMT shows higher variance ($\sigma{=}\true{0.25}$) because its constraint strength depends on token-level probabilities that vary across checkpoints.

\subsection{Cross-Model $\lambda$ Curves}
\label{sec:appendix_cross_model}

\begin{figure*}[t]
  \centering
  \includegraphics[width=\textwidth]{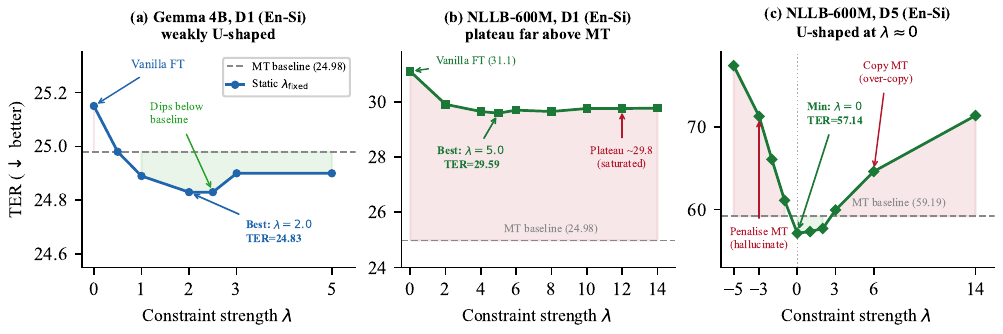}
  \caption{TER vs.\ $\lambda$ curves across model architectures. \textbf{Left:} Gemma 4B on D1 (seed 42): weakly U-shaped, the only model to dip below the MT baseline ($\Delta$TER${=}\true{+0.15}$ at $\lambda{=}\true{2.0}$). Compare with the monotonic 1B curve in Figure~\ref{fig:lambda_curves}(a). \textbf{Center:} NLLB-600M on D1 (seed 42): decreasing until $\lambda{=}5$, then plateauing at TER${\approx}\true{29.8}$ ($\lambda{=}15$ and $\lambda{=}20$ give identical results, confirming constraint saturation). Best TER~\true{29.59} at $\lambda{=}5.0$, still \true{4.61} above MT; the encoder--decoder's over-editing is too severe for an inference-time constraint to undo. \textbf{Right:} NLLB-600M on D5 (seed 42): U-shaped, centered at $\lambda{\approx}0$; neither constraint direction improves performance. Negative $\lambda$ penalizes MT-window tokens; the steeper left arm is consistent with MT structure providing useful scaffolding. The D1/D5 contrast is consistent with task differences rather than architecture alone.}
  \label{fig:lambda_scale}
\end{figure*}

Figure~\ref{fig:lambda_scale} extends the $\lambda$-curve analysis from Gemma 3 1B (Figure~\ref{fig:lambda_curves}) to the 4B scale and to NLLB-600M on both D1 and D5, showing the curves for additional architectures and task conditions.

\begin{table}[!ht]
\centering
\footnotesize
\setlength{\tabcolsep}{3pt}
\begin{tabular}{lccr}
\toprule
\textbf{Parameter} & \textbf{Gemma 1B} & \textbf{Gemma 4B} & \textbf{NLLB} \\
\midrule
Learning rate & \true{3e-5} & \true{3e-5} & \true{2e-4} \\
Scheduler & \true{Linear} & \true{Linear} & \true{Linear} \\
Warmup ratio & \true{0.10} & \true{0.10} & \true{0.10} \\
Batch (eff.) & \true{64} & \true{16$\times$4} & \true{48$\times$1} \\
Max epochs & \true{10} & \true{10} & \true{8} \\
Eval steps & \true{50} & \true{50} & \true{100} \\
Early stop. & \true{4 ($\delta$=.002)} & \true{4 ($\delta$=.002)} & \true{2 evals} \\
Weight decay & \true{0.01} & \true{0.01} & \true{0.01} \\
LoRA $r$/$\alpha$ & \true{32/32} & \true{32/32} & \true{32/64} \\
LoRA dropout & \true{0.00} & \true{0.00} & \true{0.05} \\
Quantization & None & 4-bit NF4 & None \\
Max seq.\ len. & \true{2048} & \true{2048} & \true{512/256} \\
\midrule
Time (A100) & \true{$\sim$0.5h} & \true{$\sim$0.5h} & \true{$\sim$0.5h} \\
\bottomrule
\end{tabular}
\caption{Training hyperparameters and compute. Batch size is per-device $\times$ gradient accumulation = effective. Decoder-only Gemma models are fine-tuned to generate the post-edit from a \texttt{Source: \{src\} \textbackslash n MT: \{mt\}} prompt; NLLB-600M takes \texttt{src </s> mt} as encoder input.}
\label{tab:training_details}
\end{table}

\section{Dataset Examples}
\label{sec:appendix_examples}

\subsection{Edit Type Taxonomy}
\label{sec:appendix_edit_taxonomy}

Table~\ref{tab:dataset_examples} lists representative D1 examples of each edit type.

\begin{table}[t]
\centering
\small
\renewcommand{\arraystretch}{1.8}
\begin{tabular}{p{2.8cm}p{4.0cm}}
\toprule
\textbf{Edit Type} & \textbf{Example (MT $\to$ PE)} \\
\midrule
\multicolumn{2}{l}{\textit{Stylistic (MT semantically correct):}} \\[2pt]
Loanword $\to$ native & 
\makecell[l]{\true{ස්පයිඩර් $\to$ මකුළුවා} \\
[2pt]
(\textit{``Spider'' $\to$ native term})} \\[10pt]
Register shift & \makecell[l]{\true{දනිමු $\to$ දන්නවා} \\
(\textit{literary $\to$ spoken ``know''};\\shifting between diglossic forms\\\cite{Silva2019SurveyOP})} \\[4pt]
\makecell[l]{Word order\\ (SVO$\to$SOV)} & \makecell[l]{Clause reordering to \\
Sinhala-preferred order} \\[6pt]
Retranslation & \makecell[l]{Complete rewrite,\\
same meaning} \\
\midrule
\multicolumn{2}{l}{\textit{Correction (actual MT error):}} \\[2pt]
Morphological & \makecell[l]{\true{නිදිමත $\to$ නිදිමතින්} \\
(\textit{missing instrumental -ින්})} \\[8pt]
Polysemy error & \makecell[l]{\true{ණන්කාරයා $\to$ කවුන්ට්වරයා} \\
(\textit{``creditor'' $\to$ ``count'' (title)})} \\
\bottomrule
\end{tabular}
\caption{Edit type taxonomy from manual analysis of 90 (MT, PE) pairs across D1--D3 (30 per domain). Edits span both stylistic and corrective categories with no single type dominating, particularly in D1 (Literature).}
\label{tab:edit_taxonomy}
\end{table}

\begin{table*}[!t]
\centering
\small
\begin{tabular}{lp{5.5cm}p{6.5cm}c}
\toprule
\textbf{Domain} & \textbf{Source (English)} & \textbf{Key Edit Type} & \textbf{TER} \\
\midrule
D1 (Lit.) & \true{Spider replied, ``Yes, I'll take you on an adventure...''} & \true{Loanword: ස්පයිඩර්$\to$මකුළුවා (``Spider'' $\to$ native term) + verb formality} & \true{27.3} \\
D1 (Lit.) & \true{``I didn't think,'' she sobbed.} & \true{Mistranslation: කෑගැසුවාය (``screamed'') $\to$ වැළපුණා ය (``sobbed'')} & \true{33.3} \\
D1 (Lit.) & \true{Thank you, Mrs.\ Goose.} & \true{Loanword: මිසිස් ගූස්$\to$පාත්ත මහත්මිය (English title $\to$ native)} & \true{66.7} \\
D1 (Lit.) & \true{``Ahmed?'' drowsily.} & \true{Morphological: නිදිමත$\to$නිදිමතින් (missing instrumental case -ින්)} & \true{50.0} \\
\bottomrule
\end{tabular}
\caption{Representative examples from D1, illustrating the heterogeneity of edit types. The examples include loanword replacement, mistranslation correction, and morphological fixes.}
\label{tab:dataset_examples}
\end{table*}

\subsection{LLM-Based Edit-Type Classification}
\label{sec:appendix_edit_classification}

To scale edit-type analysis beyond manual annotation, we classify \true{200} stratified samples per domain using GPT-5.2 with structured output. We adopt a simplified subset of the Multidimensional Quality Metrics (MQM) framework~\citep{lommel-etal-2014-mqm}, following the MQM-Core error typology used in WMT human evaluation campaigns~\citep{freitag-etal-2021-experts}. For each sample, we produce word-level diffs and assign one primary edit from 14 categories. Thirteen belong to four MQM groups: \textbf{Accuracy} (mistranslation, polysemy, omission, addition, untranslated), \textbf{Fluency} (morphological, agreement, punctuation), \textbf{Style} (loanword nativization, register shift, word order, lexical preference), and \textbf{Retranslation} (complete rewrite). The residual \texttt{no\_meaningful\_diff} class covers exact or near-identical copies. We classify three views per domain: human edits (MT${\to}$PE), model actions (MT${\to}$predicted), and remaining gap (predicted${\to}$PE), enabling direct comparison of human corrections and model actions.

\paragraph{D1 (Literature): Per-type breakdown.}
Table~\ref{tab:edit_types_d1} presents the full per-type classification for D1. Human edits are dominated by stylistic changes: lexical preference (\true{16.5\%}), register shift (\true{13.0\%}), word order (\true{7.0\%}), and loanword nativization (\true{4.5\%}). Among genuine errors, mistranslation is the most common (\true{22.0\%}). The model's actions are strikingly narrow: \true{71.5\%} exact copies, \true{20.0\%} punctuation changes, and \true{5.0\%} mistranslation fixes. No retranslations, no register shifts, no word order changes.

\begin{table}[t]
\centering
\small
\setlength{\tabcolsep}{4.5pt}
\begin{tabular}{lrrrc}
\toprule
\textbf{MQM Category} & \textbf{Human} & \textbf{Copied} & \textbf{Fixed} & \textbf{Copy\%} \\
\midrule
Accuracy & \true{59} & \true{48} & \true{0} & \true{81.4} \\
Fluency & \true{41} & \true{26} & \true{7} & \true{63.4} \\
Style & \true{82} & \true{56} & \true{1} & \true{68.3} \\
Retranslation & \true{17} & \true{12} & \true{0} & \true{70.6} \\
\midrule
\textbf{Total} & \true{200} & \true{143} & \true{8} & \true{71.5} \\
\bottomrule
\end{tabular}
\caption{Edit-type decomposition of Binary Collapse on D1 (Gemma 3 1B, seed~42, \true{200} samples). \textbf{Human}: number of samples where humans made this type of edit. \textbf{Copied}: model copied MT exactly. \textbf{Fixed}: model output matches PE. Binary Collapse is not selective: the model copies at high rates across all categories. Of \true{8} successful corrections, \true{7} are punctuation. (Summarized in Section~\ref{sec:binary_collapse}.)}
\label{tab:edit_decomposition}
\end{table}

\begin{table}[t]
\centering
\small
\begin{tabular}{lrrr}
\toprule
\textbf{Edit Type} & \textbf{Human} & \textbf{Model} & \textbf{Gap} \\
\midrule
\multicolumn{4}{l}{\textit{Accuracy}} \\
\hspace{1em}mistranslation & \true{44} & \true{10} & \true{45} \\
\hspace{1em}addition & \true{6} & -- & \true{6} \\
\hspace{1em}untranslated & \true{4} & \true{2} & \true{3} \\
\hspace{1em}polysemy\_sense & \true{3} & -- & \true{1} \\
\hspace{1em}omission & \true{2} & \true{1} & \true{2} \\
\midrule
\multicolumn{4}{l}{\textit{Fluency}} \\
\hspace{1em}punctuation & \true{29} & \true{40} & \true{25} \\
\hspace{1em}morphological & \true{12} & -- & \true{15} \\
\midrule
\multicolumn{4}{l}{\textit{Style}} \\
\hspace{1em}lexical\_preference & \true{33} & \true{1} & \true{39} \\
\hspace{1em}register\_shift & \true{26} & \true{1} & \true{24} \\
\hspace{1em}word\_order & \true{14} & -- & \true{12} \\
\hspace{1em}loanword\_nativ. & \true{9} & \true{2} & \true{3} \\
\midrule
retranslation & \true{17} & -- & \true{16} \\
no\_meaningful\_diff & \true{1} & \true{143} & \true{9} \\
\bottomrule
\end{tabular}
\caption{Per-type edit classification for D1 (\true{200} samples, Gemma 3 1B, seed~42). \textbf{Human}: MT${\to}$PE edits. \textbf{Model}: MT${\to}$predicted actions. \textbf{Gap}: predicted${\to}$PE remaining differences. The model's repertoire is limited to punctuation (\true{40}) and a few mistranslation fixes (\true{10}); all other edit types are effectively abandoned.}
\label{tab:edit_types_d1}
\end{table}

\paragraph{Cross-tabulation.} Table~\ref{tab:edit_cross_d1} shows what the model does for each human edit category. When humans make style edits, the model either copies (\true{56/82}) or makes unrelated fluency changes (\true{19/82}), almost never producing actual style edits (\true{1/82}). When humans fix accuracy errors, the model copies \true{48/59} and attempts only \true{11}, with \true{0} matching PE exactly.

\begin{table}[t]
\centering
\small
\begin{tabular}{lrrrrr}
\toprule
& \multicolumn{5}{c}{\textbf{Model Action Category}} \\
\cmidrule(lr){2-6}
\textbf{Human} & \textbf{None} & \textbf{Acc.} & \textbf{Flu.} & \textbf{Sty.} & \textbf{Ret.} \\
\midrule
Accuracy & \true{48} & \true{5} & \true{4} & \true{2} & -- \\
Fluency & \true{26} & -- & \true{15} & -- & -- \\
Style & \true{56} & \true{6} & \true{19} & \true{1} & -- \\
Retranslation & \true{12} & \true{2} & \true{2} & \true{1} & -- \\
\bottomrule
\end{tabular}
\caption{Cross-tabulation of human edit category vs.\ model action category on D1 (\true{200} samples). Rows: what humans edited. Columns: what the model did. The ``None'' column (exact copies) dominates every row. When the model does act on style edits, it typically makes fluency (punctuation) changes instead.}
\label{tab:edit_cross_d1}
\end{table}

\paragraph{Cross-domain comparison.}
Table~\ref{tab:edit_cross_domain} summarizes the edit-type profile and model behavior across D1--D3, which share the same MT system (Google Translate) and task structure. The pattern is consistent: higher proportions of subjective edits correlate with lower model success rates and higher copy rates. The net edit effect (Table~\ref{tab:edit_behavior}) is consistent with signal \emph{inconsistency}, rather than composition, contributing to the outcome: D3 matches D1's subjective proportion yet its edits stay net-positive, while only D1's editing is net-negative.

\begin{table}[t]
\centering
\small
\begin{tabular}{lrrr}
\toprule
& \textbf{D2} & \textbf{D3} & \textbf{D1} \\
& \textit{News} & \textit{Wiki} & \textit{Lit.} \\
\midrule
\multicolumn{4}{l}{\textit{Human edit distribution (\%)}} \\
\hspace{1em}Accuracy & \true{7.5} & \true{19.0} & \true{29.5} \\
\hspace{1em}Fluency & \true{68.0} & \true{33.0} & \true{20.5} \\
\hspace{1em}Style & \true{19.0} & \true{41.0} & \true{41.0} \\
\hspace{1em}Retranslation & \true{5.5} & \true{7.0} & \true{8.5} \\
\midrule
\multicolumn{4}{l}{\textit{Model behavior (\%)}} \\
\hspace{1em}Copy rate & \true{40.5} & \true{51.5} & \true{71.5} \\
\hspace{1em}Success rate & \true{24.0} & \true{7.5} & \true{4.0} \\
\hspace{1em}Subjective \% & \true{24.5} & \true{48.0} & \true{49.5} \\
\bottomrule
\end{tabular}
\caption{Cross-domain edit-type profiles and model behavior (\true{200} samples per domain, Gemma 3 1B, seed~42). Subjective \% = style + retranslation. Domains ordered by increasing subjectivity. Success rate = model output exactly matches PE.}
\label{tab:edit_cross_domain}
\end{table}

D2 (News) is the cleanest domain: \true{68.0\%} of human edits are fluency corrections (mostly punctuation, \true{58.5\%} of all edits), only \true{24.5\%} are subjective, and the model achieves the highest success rate (\true{24.0\%}, \true{48/200}). D3 (Wikipedia) resembles D1 in edit profile (\true{48.0\%} subjective) but with more fluency edits (\true{33.0\%} vs \true{20.5\%}), yielding a moderate success rate (\true{7.5\%}). On D1 (Literature), \true{49.5\%} of edits are subjective, the copy rate peaks at \true{71.5\%}, and only \true{4.0\%} of model outputs match PE.

\paragraph{Editing is net-negative only on D1.}
The copy and success rates above are measured on the \true{200} stratified samples (edit-needed rows only). Table~\ref{tab:edit_behavior} reports the complementary picture on the \emph{full} test sets, including the decisive signal: the net effect of the model's edits. Over the sentences the model chooses to edit, we compute the mean change in TER relative to leaving the MT unchanged. Editing is net-negative only on D1 ($+$\true{6.3}); on D2, D3, and the low-quality-MT contrast D5 the model's edits are net-positive, even though D3's edit-type composition matches D1's (Table~\ref{tab:edit_cross_domain}). Test-time edit behavior thus tracks the consistency of the post-edit signal, paralleling the vanilla-FT degradation pattern (D1 alone degrades) and the constrained-decoding curves (Section~\ref{sec:constrained}).

\begin{table}[t]
\centering
\small
\begin{tabular}{lrrrr}
\toprule
\textbf{Domain} & \textbf{$n$} & \textbf{Edit} & \textbf{Edits} & \textbf{per-edit} \\
 & & \textbf{rate} & \textbf{$\neq$PE} & \textbf{$\Delta$TER} \\
\midrule
D1 Literature      & \true{716}  & \true{23.6} & \true{88.8} & \true{$+$6.3} \\
D2 News            & \true{1664} & \true{37.7} & \true{72.6} & \true{$-$0.9} \\
D3 Wikipedia       & \true{2184} & \true{41.6} & \true{80.6} & \true{$-$1.1} \\
\midrule
D5 (low-qual.\ MT) & \true{1292} & \true{79.8} & \true{98.4} & \true{$-$3.2} \\
\bottomrule
\end{tabular}
\caption{Test-time edit behavior on the \emph{full} test sets (Gemma 3 1B, seed~42). \textbf{Edit rate}: \% of sentences where the output differs from the MT. \textbf{Edits$\neq$PE}: of those, \% not matching the post-edit. \textbf{per-edit $\Delta$TER}: mean $[\text{TER(output,PE)} - \text{TER(MT,PE)}]$ over edited sentences; positive means editing moves the output \emph{away} from the post-edit. Editing is net-negative only on D1, the domain with the most inconsistent post-edit signal; D1, D2, D3 share the same MT (Google Translate), D5 uses NLLB and is shown as a contrast. TER computed with Asian-aware tokenization.}
\label{tab:edit_behavior}
\end{table}

\paragraph{NLLB-600M comparison.}
Applying the same analysis to NLLB-600M predictions on D1 reveals a different failure mode. The encoder--decoder copies less (\true{48.0\%} vs Gemma's \true{71.5\%}), editing more aggressively, yet achieves an even lower success rate (\true{2.5\%} vs \true{4.0\%}). Copy rates are uniformly lower across all categories (accuracy \true{52.6\%}, style \true{50.6\%}, fluency \true{34.1\%}), confirming that NLLB over-edits indiscriminately rather than selectively. The model's extra edits land mostly in fluency (punctuation: \true{61/104} non-copy actions), mirroring Gemma's narrow repertoire at higher volume. This is consistent with NLLB's catastrophic TER degradation (\true{$-$5.80}; Section~\ref{sec:miscalibration}): the over-editing is too severe for inference-time correction.

\paragraph{Classification prompt.}
Each batch of 10 samples is sent to GPT-5.2 with structured output (Pydantic schema) at temperature~0. The system prompt instructs the model to act as a Sinhala linguistics expert and provides the full taxonomy with Sinhala-specific examples. The user message supplies the English source, word-level diff, and TER for each sample. The full system prompt is shown below.

\begin{quote}
\small
\texttt{You are a Sinhala linguistics expert classifying edit types in English-to-Sinhala translation post-editing.}

\medskip
\texttt{MQM-Core Edit Type Taxonomy (adapted for MT post-editing):}

\medskip
\texttt{Category 1 --- ACCURACY errors (genuine MT errors):}\\
\texttt{~~mistranslation, polysemy\_sense, omission, addition, untranslated}

\medskip
\texttt{Category 2 --- FLUENCY errors (semantically correct, grammatically ill-formed):}\\
\texttt{~~morphological, agreement, punctuation}

\medskip
\texttt{Category 3 --- STYLE/REGISTER preferences (correct and well-formed):}\\
\texttt{~~loanword\_nativization, register\_shift, word\_order, lexical\_preference}

\medskip
\texttt{Category 4 --- RETRANSLATION:}\\
\texttt{~~retranslation (complete rewrite, semantically adequate)}

\medskip
\texttt{Other: no\_meaningful\_diff (for [EXACT COPY] or near-identical)}

\medskip
\texttt{For each sample: (1) classify PRIMARY edit type, (2) list secondary types, (3) set mqm\_category, (4) write 5--10 word note.}
\end{quote}

\noindent Each sample in the user message follows the format: \texttt{id}, \texttt{SRC} (English source), \texttt{diff} (word-level changes marked with \texttt{(- deleted -)} and \texttt{(+ inserted +)}), and \texttt{TER}. The structured output schema enforces a list of classifications, each with \texttt{primary\_type} (enum), \texttt{secondary\_types} (list of enums), \texttt{mqm\_category} (string), and \texttt{brief\_note} (string).

\subsection{Qualitative Model Output Examples}
\label{sec:appendix_qualitative}

Table~\ref{tab:qualitative} presents three qualitative examples from D1, contrasting the vanilla fine-tuned model with the static constraint.

\begin{table*}[!t]
\centering
\small
\begin{tabular}{p{1.5cm}p{13cm}}
\toprule
\multicolumn{2}{l}{\textbf{Example 1: Unnecessary Edit} (MT $=$ PE, TER\textsubscript{ref} $= 0$)} \\
\midrule
Source & \true{``I must buy you.''} \\
MT Output & \true{``මම ඔබව මිලදී ගත යුතුයි.''} \\
Human PE & \true{``මම ඔබව මිලදී ගත යුතුයි.''} (identical to MT) \\
Vanilla FT & \true{``මම ඔබව මිලදී {\color{editcol}ගන්නවා}.''} (changed \true{ගත යුතුයි} ``must buy'' $\to$ \true{ගන්නවා} ``will buy'': unnecessary modality shift) \\
Constrained & \true{``මම ඔබව මිලදී ගත යුතුයි.''} (correctly preserved MT, TER $= 0$) \\
\midrule
\multicolumn{2}{l}{\textit{MT is correct. The model changes obligation to simple future at high confidence, altering the meaning.}} \\
\toprule
\multicolumn{2}{l}{\textbf{Example 2: Stylistic Over-Edit} (PE makes 2 small changes; model corrupts a content word)} \\
\midrule
Source & \true{Cut off a cobra's head and it could only wriggle until sunset.} \\
MT Output & \true{නාගයෙකුගේ හිස කපා දමන්න, එය ඉර බැස යන තුරු පමණක් දඟලන්නට විය.} \\
Human PE & \true{නාගයෙකුගේ හිස කපා දමන්න, {\color{editcol}ඌට} ඉර බැස යන තුරු පමණක් දඟලන්නට {\color{editcol}පුළුවන}.} \\
& (\textit{Only 2 changes: pronoun} \true{එය}$\to$\true{ඌට} \textit{and verb form} \true{විය}$\to$\true{පුළුවන}) \\
Vanilla FT & \true{{\color{editcol}නැගියෙකුගේ} හිස කපා දමන්න, එය ඉර බැස යන තුරු පමණක් දඟලන්නට විය.} (corrupts \true{නාගයෙකුගේ} ``cobra's'' $\to$ \true{නැගියෙකුගේ}, a non-word) \\
Constrained & \true{නාගයෙකුගේ හිස කපා දමන්න, එය ඉර බැස යන තුරු පමණක් දඟලන්නට විය.} (preserves MT) \\
\midrule
\multicolumn{2}{l}{\textit{PE makes minor stylistic adjustments. The model ignores PE's changes, instead corrupting ``cobra'' into a non-word.}} \\
\toprule
\multicolumn{2}{l}{\textbf{Example 3: Correct Intent, Wrong Execution} (MT leaves proper noun in Latin script)} \\
\midrule
Source & \true{That weekend, she went to Ville Belle with her father.} \\
MT Output & \true{එම සති අන්තයේ ඇය තම පියා සමඟ Ville Belle වෙත ගියාය.} (\true{Ville Belle} left untransliterated) \\
Human PE & \true{එම සති අන්තයේ, ඇය තම පියා සමඟ {\color{editcol}විල්} {\color{editcol}බෙල්} වෙත ගියාය.} (correctly transliterates to \true{විල් බෙල්}) \\
Vanilla FT & \true{එම සති අන්තයේ, ඇය තම පියා සමඟ {\color{editcol}විල්} {\color{editcol}බල්ෙ} වෙත ගියාය.} (attempts transliteration but produces \true{විල් බල්ෙ}, an incorrect rendering) \\
Constrained & \true{එම සති අන්තයේ ඇය තම පියා සමඟ Ville Belle වෙත ගියාය.} (preserves MT) \\
\midrule
\multicolumn{2}{l}{\textit{Both PE and vanilla identify the same issue (untransliterated name), but the model hallucinates the wrong transliteration.}} \\
\bottomrule
\end{tabular}
\caption{Qualitative examples from D1 (all text is real model output from Gemma 3 1B, seed~42; constrained $= \lambda_{\text{static}}$ at $\lambda{=}2.0$). Example~1: the model edits perfect MT. Example~2: the model ignores PE's minor stylistic fixes and instead corrupts a content word. Example~3: the model attempts the same correction as PE but hallucinates the wrong transliteration. Constrained decoding preserves the MT output in all three cases.}
\label{tab:qualitative}
\end{table*}

Example~2 is the prototypical case: the reference PE changes only a pronoun form and a verb ending (both stylistic), yet the vanilla model, rather than replicating these minor fixes, corrupts a content word (``cobra'') into a non-word. Example~3 shows a rarer case where both PE and the model identify the same genuine error (an untransliterated proper noun), but the model hallucinates the wrong transliteration. In both cases, the entropy-based constraint would \emph{not} intervene (the model is confident in its edits), which is precisely why it underperforms the static constraint.

\section{MQM Classification Validation}
\label{sec:appendix_mqm_validation}

\subsection{Annotator Instructions}

Three native Sinhala speakers with professional or near-professional bilingual competence in English and Sinhala were the annotators. Annotators were compensated at fair market rates and completed the task independently with no inter-annotator discussion.

The following instructions were provided in both English and Sinhala:

\begin{quote}
\small\ttfamily
You will be shown an English sentence, its Sinhala machine translation (MT), the human post-edited version, and a word-level diff showing what changed between MT and post-edit. An automated system has assigned a category label to the edit. Your task is to judge whether this label correctly describes the PRIMARY change made by the post-editor.

The four categories are:
\begin{itemize}
    \item ACCURACY: The MT contained a genuine error (wrong meaning, mistranslation, omission, addition, untranslated text). The post-editor fixed this error.
    \item FLUENCY: The MT was semantically correct but grammatically ill-formed (wrong morphological form, agreement error, punctuation). The post-editor fixed the grammatical issue.
    \item STYLE/REGISTER: The MT was semantically correct and grammatically well-formed, but the post-editor changed it for stylistic reasons (replaced a loanword with a native term, changed register from formal to colloquial, reordered words, substituted a synonym).
    \item RETRANSLATION: The post-editor completely rewrote the sentence while preserving the meaning.
\end{itemize}

For each sample answer: (1) Is the assigned label correct? YES / NO / PARTIALLY. (2) If NO or PARTIALLY, select the correct category.
\end{quote}

\subsection{Calibration Examples}

Before the main task, annotators completed five calibration samples with correct answers revealed. Two examples are shown in Table~\ref{tab:calibration_examples} to illustrate the ACCURACY/STYLE distinction, which is the most frequently confounded boundary.

\begin{table}[!ht]
\centering
\small
\resizebox{\columnwidth}{!}{%
\begin{tabular}{p{1.2cm}p{5.0cm}p{1.8cm}}
\toprule
\textbf{Field} & \textbf{Content} & \textbf{Label} \\
\midrule
Source & ``I didn't think,'' she sobbed. & \\
MT & \true{``මම හිතුවේ නැහැ,'' ඇය {\color{editcol}කෑගැසුවාය}.} & \\
PE & \true{``මම හිතුවේ නැහැ,'' ඇය {\color{editcol}වැළපුණා} {\color{editcol}ය}.} & \\
Diff & \true{කෑගැසුවාය} (screamed) $\to$ \true{වැළපුණා ය} (sobbed) & \textbf{ACCURACY} \\
\midrule
Source & Spider replied, ``Yes...'' & \\
MT & \true{{\color{editcol}ස්පයිඩර්} පිළිතුරු දුන්නේ...} & \\
PE & \true{{\color{editcol}මකුළුවා} පිළිතුරු දුන්නේ...} & \\
Diff & \true{ස්පයිඩර්} (loanword) $\to$ \true{මකුළුවා} (native) & \textbf{STYLE} \\
\bottomrule
\end{tabular}}
\caption{Calibration examples for the ACCURACY/STYLE boundary. The first is ACCURACY (``screamed'' is a mistranslation of ``sobbed''). The second is STYLE (both are valid Sinhala renderings; the native term is a stylistic preference).}
\label{tab:calibration_examples}
\end{table}

\subsection{Sample Selection}

We drew \true{100} samples from the GPT-5.2 classified D1 set, stratified by MQM category: \true{25} each for Accuracy, Fluency, and Style; all \true{17} available Retranslation samples; and \true{8} additional samples drawn proportionally from the larger categories to reach \true{100}. Samples within each category were selected randomly (seed 42). The same \true{100} samples were evaluated by all three annotators.

\subsection{Results}

Table~\ref{tab:mqm_validation} reports per-category agreement and Fleiss' $\kappa$.

\begin{table}[!ht]
\centering
\small
\resizebox{\columnwidth}{!}{%
\begin{tabular}{lcccc}
\toprule
\textbf{Category} & \textbf{GPT (\%)} & \textbf{Corrected (\%)} & \textbf{Agree (\%)} & \textbf{$\kappa$} \\
\midrule
Accuracy & \true{26} & \true{29} & \true{73} & \true{0.64} \\
Fluency & \true{26} & \true{27} & \true{77} & \true{0.72} \\
Style & \true{31} & \true{28} & \true{65} & \true{0.62} \\
Retranslation & \true{17} & \true{16} & \true{82} & \true{0.92} \\
\midrule
\textbf{Overall} & \true{100} & \true{100} & \true{73} & \true{0.70} \\
\bottomrule
\end{tabular}}
\caption{GPT-5.2 classification validation. \textbf{GPT}: original automated distribution. \textbf{Corrected}: distribution after human corrections. \textbf{Agree}: proportion confirmed by all three annotators. $\kappa$: per-category binary Fleiss' $\kappa$ across the three annotators (overall: multiclass Fleiss' $\kappa$).}
\label{tab:mqm_validation}
\end{table}

\subsection{Confusion Matrix}

Table~\ref{tab:mqm_confusion} shows the confusion matrix between GPT-5.2 labels and consensus human labels.

\begin{table}[!ht]
\centering
\small
\begin{tabular}{lcccc}
\toprule
\textbf{GPT label $\to$} & \textbf{Acc.} & \textbf{Flu.} & \textbf{Sty.} & \textbf{Ret.} \\
\midrule
\textbf{Human: Accuracy} & \true{26} & \true{0} & \true{3} & \true{0} \\
\textbf{Human: Fluency} & \true{0} & \true{26} & \true{1} & \true{0} \\
\textbf{Human: Style} & \true{0} & \true{0} & \true{27} & \true{1} \\
\textbf{Human: Retranslation} & \true{0} & \true{0} & \true{0} & \true{16} \\
\bottomrule
\end{tabular}
\caption{Confusion matrix between GPT-5.2 labels (columns) and consensus human labels (rows) on \true{100} validation samples. Diagonal entries indicate agreement; off-diagonal entries indicate systematic confusions. GPT overclassifies Style: \true{3} cases relabeled as Accuracy, \true{1} as Fluency.}
\label{tab:mqm_confusion}
\end{table}

\section{Evaluation Metrics}
\label{sec:appendix_metrics}

All TER scores use SacreTER~\citep{post-2018-call} with default tokenization (no normalization); BLEU uses SacreBLEU with \texttt{intl} tokenization; chrF++ uses $\beta{=}2$ with word bigrams (\texttt{word\_order}${=}2$). The MT baseline is evaluated under identical settings. Zero-shot TER values for base LLMs are high because uninstructed models generate fluent but task-unrelated Sinhala text (Appendix~\ref{sec:appendix_prompts}).

\subsection{Metric Robustness of the Diagnostic}
\label{sec:appendix_metric_robustness}

The diagnostic reads the shape of the constraint curve, not absolute
scores, so it should not depend on the metric that traces the curve.
We recomputed the static-constraint sweep under chrF++ and BLEU for D1,
D5, and English--Marathi (Gemma 3 1B, seed 42).
Table~\ref{tab:metric_robustness} reports the best improvement over
the MT baseline under each metric. The U-shaped versus monotonic
distinction holds under every metric: on D1 the gains are within
noise on all three, while on D5 and English--Marathi the curve peaks
clearly above the MT baseline on each. The calibration finding
(Table~\ref{tab:confidence}) is measured on the probability and
entropy values directly, so it is independent of the sequence metric.

\begin{table}[!ht]
\centering
\small
\begin{tabular}{lccc}
\toprule
\textbf{Dataset} & \textbf{$\Delta$TER} & \textbf{$\Delta$chrF++} & \textbf{$\Delta$BLEU} \\
\midrule
D1 (monotonic) & \true{+0.03} & \true{+0.07} & \true{+0.05} \\
D5 (U-shaped) & \true{+3.98} & \true{+0.97} & \true{+2.58} \\
En-Mr (U-shaped) & \true{+1.20} & \true{+0.90} & \true{+3.03} \\
\bottomrule
\end{tabular}
\caption{Best improvement over the MT baseline along the static
$\lambda$ sweep, per metric (Gemma 3 1B, seed 42). All deltas are
oriented so that positive is better. The monotonic/U-shaped
distinction survives under chrF++ and BLEU.}
\label{tab:metric_robustness}
\end{table}

\section{Failed Interventions}
\label{sec:appendix_failed}

Before arriving at constrained decoding as a diagnostic, we explored three training-time interventions aimed at reducing the over-editing behavior documented in Section~\ref{sec:binary_collapse}. Each targets a different hypothesis about the failure mechanism: \textbf{Hidden State Anchoring (HSA)} regularizes the model's internal representation of the MT span to prevent drift during fine-tuning; \textbf{Edit-Level Conditioning (ELC)} provides explicit edit-level tags (none/minor/major) as input to guide the model's editing intensity; and \textbf{Chain-of-Thought (CoT) Self-Tagging} trains the model to predict its own edit level before generating the post-edit, testing whether the model can learn to self-diagnose. All three failed to reduce over-editing, and in some cases worsened TER substantially (Table~\ref{tab:failed}). In these experiments, the failures show that the tested internal signals did not provide reliable regularization targets (HSA), conditioning signals (ELC), or self-predicted tags (CoT), motivating the inference-time, confidence-agnostic constraints in Section~\ref{sec:constrained}.

\paragraph{Hidden State Anchoring (HSA).} HSA adds an auxiliary regularization loss during training that penalizes drift in the model's hidden-state representation of the MT span. At each training step: (1)~identify the MT token span in the input, (2)~extract hidden states from the final Transformer layer and apply layer normalization, (3)~mean-pool the MT span hidden states into a single vector $\mathbf{h}_{\text{mt}}$, and (4)~add a weighted MSE penalty:
\begin{equation}
    \mathcal{L}_{\text{total}} = \mathcal{L}_{\text{APE}} + \alpha \cdot \|\mathbf{h}_{\text{mt}} - \mathbf{a}\|^2
\label{eq:hsa}
\end{equation}
where $\mathbf{a}$ is the anchor target. We ablated three anchor strategies (Section~\ref{sec:appendix_hsa}): \textbf{zero-scalar} ($\alpha{=}0.1$), \textbf{zero-vector} ($\alpha{=}0.001$), and \textbf{MT-vector} ($\alpha{=}0.001$). The simplest variant (zero-scalar) performed best, but with high seed variance ($\sigma{=}\true{0.26}$).

\paragraph{Edit-Level Conditioning (ELC).} The training prompt augments the input with a gold edit-level tag:

\begin{quote}
\small\ttfamily
Source: \{source\}\\
MT: \{mt\}\\
Edit level: \{none $|$ minor $|$ major\}
\end{quote}

\noindent Tags are derived from TER(MT, PE): TER${=}0$ $\to$ \texttt{none}, TER${<}20$ $\to$ \texttt{minor}, TER${\geq}20$ $\to$ \texttt{major}. At inference, gold tags are unavailable (they require the reference), so a single fixed tag must be applied to all test samples. ``ELC (oracle)'' uses per-sample gold tags (a cheating upper bound) but actually \emph{degrades} performance (TER~\true{27.29} vs.\ baseline~\true{24.98}), primarily due to noise on the \texttt{none} class where the model still hallucinates edits. The fixed \texttt{minor} tag (TER~\true{25.88}) performs comparably to vanilla fine-tuning, confirming that a single tag cannot capture the heterogeneous edit requirements. This is consistent with the QE results in Appendix~\ref{sec:appendix_qe}, where even fine-tuned XLM-R achieves only AUROC~\true{0.745} for edit-level detection.

\paragraph{Chain-of-Thought (CoT) Self-Tagging.} Instead of providing the edit level as input, the model is trained to \emph{predict} it before generating the post-edit. The input contains only source and MT:

\begin{quote}
\small\ttfamily
Source: \{source\}\\
MT: \{mt\}
\end{quote}

\noindent and the target is the edit-level tag followed by the post-edited text:

\begin{quote}
\small\ttfamily
[none $|$ minor $|$ major]\\
\{post-edited text\}
\end{quote}

\noindent Table~\ref{tab:class_collapse} shows that the model's predicted tag distribution diverges dramatically from the gold distribution. The confusion matrix confirms: \true{90\%} of gold-minor samples (135/150) are tagged as \texttt{[major]}, and \true{66\%} of gold-none samples (137/208) are also tagged as \texttt{[major]}. The resulting TER (\true{29.55}) is the worst of all methods, with the heaviest degradation on the none class ($\Delta$TER${=}\true{-7.09}$).

\begin{table}[!ht]
\centering
\small
\begin{tabular}{lccc}
\toprule
\textbf{Tag} & \textbf{Gold (\%)} & \textbf{Predicted (\%)} & \textbf{$\Delta$} \\
\midrule
\texttt{none} & \true{29.1} & \true{17.7} & \true{$-$11.4} \\
\texttt{minor} & \true{20.9} & \true{0.1} & \true{$-$20.8} \\
\texttt{major} & \true{50.0} & \true{82.1} & \true{$+$32.1} \\
\midrule
Tag accuracy & \multicolumn{3}{c}{\true{53.9\%}} \\
\bottomrule
\end{tabular}
\caption{CoT self-tagging: gold vs.\ predicted edit-level distribution on D1 test set. The minor class is virtually eliminated ($-$20.8pp), absorbed almost entirely by the major class ($+$32.1pp).}
\label{tab:class_collapse}
\end{table}

\subsection{Summary of Failed Interventions}

\begin{table}[!ht]
\centering
\small
\begin{tabular}{lcccc}
\toprule
\textbf{Method} & \textbf{TER} $\downarrow$ & \textbf{$\Delta$TER} & \textbf{$\sigma$} & \textbf{Note} \\
\midrule
MT Baseline & \true{24.98} & -- & -- & -- \\
Vanilla FT & \true{25.84} & \true{$-$0.86} & \true{0.02} & -- \\
\midrule
HSA (mean) & \true{25.58} & \true{$-$0.60} & \true{0.26} & High $\sigma$ \\
\midrule
ELC (oracle) & \true{27.29} & \true{$-$2.31} & -- & \makecell{Noise on \\ \texttt{none} class} \\
ELC (minor) & \true{25.88} & \true{$-$0.90} & -- & $\approx$ vanilla \\
\midrule
CoT & \true{29.55} & \true{$-$4.57} & -- & \makecell{\true{82\%} predict \\ \texttt{[major]}} \\
\bottomrule
\end{tabular}
\caption{Failed interventions on D1 (Literature), Gemma 3 1B. HSA is seed-dependent ($\sigma{=}\true{0.26}$); ELC degrades even with oracle tags; CoT collapses to predicting \texttt{[major]} for \true{82\%} of samples.}
\label{tab:failed}
\end{table}

\section{HSA Ablation}
\label{sec:appendix_hsa}

Table~\ref{tab:hsa_ablation} compares the three HSA anchor strategies on D1 (Gemma 3 1B, seed 42). All variants use the final Transformer layer with layer normalization; $\alpha$ was tuned per variant: $\alpha{=}0.1$ for zero-scalar and $\alpha{=}0.001$ for both zero-vector and MT-vector.

\begin{table}[!ht]
\centering
\footnotesize
\setlength{\tabcolsep}{3pt}
\begin{tabular}{llccc}
\toprule
\textbf{Anchor} & \textbf{Target} & $\boldsymbol{\alpha}$ & \textbf{TER}$\downarrow$ & \textbf{$\Delta$} \\
\midrule
Vanilla FT & -- & -- & \true{25.84} & -- \\
\midrule
Zero-scalar & $\bar{h}_{\text{mt}} {\to} 0$ & 0.1 & \true{25.26} & \true{+0.58} \\
MT-vector & $\mathbf{h} {\to} \mathbf{h}^{\text{frz}}$ & 0.001 & \true{25.73} & \true{+0.11} \\
Zero-vector & $\mathbf{h} {\to} \mathbf{0}$ & 0.001 & \true{25.69} & \true{+0.15} \\
\bottomrule
\end{tabular}
\caption{HSA anchor ablation on D1 (seed 42). Zero-scalar performs best but is modest ($+$0.58 TER) and highly seed-dependent ($\sigma{=}\true{0.26}$).}
\label{tab:hsa_ablation}
\end{table}

\noindent \textbf{Zero-scalar} penalizes the global mean of all pooled MT span features toward zero, a coarse regularizer that prevents the model from drifting the MT representation space too far from initialization. \textbf{MT-vector} anchors each sample's pooled MT hidden state to the corresponding representation from the frozen base model (precomputed before fine-tuning). \textbf{Zero-vector} anchors each sample directly to the zero vector via per-sample MSE. Paradoxically, the simplest variant (zero-scalar) outperforms the more targeted alternatives, suggesting that the fine-grained per-sample anchoring adds noise in the low-resource setting.

\section{Constrained Decoding Details}
\label{sec:appendix_decoding}

\subsection{Algorithm}

Algorithm~1 presents the pseudocode for constrained decoding. At each step, the method computes a (possibly dynamic) $\lambda_t$, then subtracts it from all token logits and adds it back for tokens in the MT alignment window $\mathcal{W}_t$. A pointer tracks alignment through the MT sequence and advances when the generated token matches a window token. When no window token matches (e.g., under reordering or insertion), the pointer does not advance and the window effectively widens until alignment is re-established. This makes the scheme robust to local reordering, at the cost of occasionally under-penalizing divergent continuations.

\begin{table}[!ht]
\centering
\small
\begin{tabular}{p{7cm}}
\toprule
\textbf{Algorithm 1:} Constrained Decoding \\
\midrule
\textbf{Input:} Model $M$, MT tokens $\mathbf{x}^{\text{mt}}$, type, $\lambda$, window $k$, $\gamma$ or $\beta$ \\
\textbf{Output:} Post-edited tokens $\mathbf{y}$ \\
\midrule
Initialize MT pointer $\text{ptr} \gets 0$ \\
\textbf{for} $t = 1, 2, \ldots$ \textbf{do} \\
\quad $\mathbf{l}_t \gets M.\text{logits}(\mathbf{y}_{<t})$ \\
\quad $p_t \gets \text{softmax}(\mathbf{l}_t)$ \\
\quad $\mathcal{W}_t \gets \{\, j : |\,j - \text{ptr}\,| \leq k \,\}$ \\
\quad $\lambda_t \gets$ \textsc{GetLambda}($p_t$, $\mathcal{W}_t$, type, $\lambda$) \\
\quad $\mathbf{l}_t[\,:\,] \mathrel{-}= \lambda_t$ \quad\textit{// penalize all tokens} \\
\quad $\mathbf{l}_t[\{x_j^{\text{mt}} : j \in \mathcal{W}_t\}] \mathrel{+}= \lambda_t$ \quad\textit{// restore MT-window tokens} \\
\quad $y_t \gets \text{decode}(\mathbf{l}_t)$ \\
\quad Update ptr based on $y_t$ and $\mathcal{W}_t$ \\
\textbf{end for} \\
\midrule
\textsc{GetLambda}($p_t$, $\mathcal{W}_t$, type, $\lambda$): \\
\quad \textbf{if} type $=$ static: \textbf{return} $\lambda$ \\
\quad \textbf{if} type $=$ entropy: \textbf{return} $\lambda \cdot (H(p_t)/H_{\max})^\gamma$ \\
\quad \textbf{if} type $=$ prob: \textbf{return} $\lambda \cdot (\max_{j \in \mathcal{W}_t} p_t(x_j^{\text{mt}}))^\beta$ \\
\bottomrule
\end{tabular}
\label{alg:constrained}
\end{table}

\subsection{Hyperparameter Search}

\begin{table}[!ht]
\centering
\small
\resizebox{\columnwidth}{!}{%
\begin{tabular}{lcc}
\toprule
\textbf{Parameter} & \textbf{Search Range} & \textbf{Optimal} \\
\midrule
$\lambda$ (all) & 0.1, 0.3, 0.5, 1.0, 2.0, 2.5, 3.0 & \true{2.5} \\
$\gamma$ (entropy) & 0.5, 1.0, 2.0, 3.0 & \true{2.0} \\
$\beta$ (probability) & 0.5, 1.0, 2.0 & \true{1.0} \\
$k$ (window) & 1, 2, 3 & \true{2} \\
\bottomrule
\end{tabular}}
\caption{Hyperparameter search ranges and optimal values (tuned on D1 validation set, Gemma 3 1B).}
\label{tab:hyperparams}
\end{table}

\subsection{The Sequence-Level Variant}
\label{sec:appendix_seqnll}

The sequence-scaled variant ($\lambda_{\text{seq-NLL}}$,
Table~\ref{tab:constrained}) replaces the per-token confidence
signal with one computed over the whole generated prefix. At step
$t$, let $c_t$ be the running geometric-mean token probability of
the tokens generated so far, i.e.\ the exponentiated
length-normalized log-likelihood of the prefix. The base penalty is
scaled by the sequence-level \emph{uncertainty} $(1 - c_t)$:
\begin{equation}
\lambda_t = \lambda \cdot (1 - c_t)^{\gamma}, \;
c_t = \exp\!\bigl(\tfrac{1}{t-1}\textstyle\sum_{s < t} \log p_s(y_s)\bigr),
\label{eq:seqnll}
\end{equation}
with $\gamma{=}\true{1}$, so the constraint relaxes when the model
is confident about the sequence as a whole and tightens when it is
not, mirroring the entropy variant at the sequence level. Because
the fine-tuned model is overconfident even at the sequence level
($c_t \approx \true{0.94}$ throughout, as most tokens are
high-probability copies), the scaling factor $(1 - c_t)$ is small,
and the base $\lambda$ grid is widened accordingly to
$\{\true{0, 2, 5, 10, 20, 40, 60}\}$ so that the effective penalty
can reach the static optimum. On D1 (Gemma 3 1B, 3 seeds), its
best-$\lambda$ TER is \true{25.32$\pm$0.21}: between the two
token-level variants and, like them, short of the confidence-free
static constraint (\true{24.98$\pm$0.08}). The Static $>$
confidence-aware ordering therefore does not depend on reading
confidence one token at a time.

\subsection{Sensitivity to the Window Size and Penalty Parameters}
\label{sec:appendix_ksens}

The diagnostic's conclusions could in principle depend on the
window size $k$ and the penalty parameters $\gamma$ and $\beta$. They
do not. $\lambda$ itself is swept rather than tuned, so no single
value produces the curve shape, and the static constraint that
carries the shape signal has no $\gamma$ or $\beta$; those parameters
only affect the two dynamic variants.

For $k$, we reran the static sweep on D1 at $k{=}1$ and $k{=}3$
(Gemma 3 1B, seed 42). The curve stays monotonically decreasing at
every $k$, and its lowest point recovers the MT baseline
(TER \true{24.98}): best-$\lambda$ TER is \true{24.98},
\true{24.96}, and \true{25.09} for $k{=}1$, $2$, $3$. The window
size changes how sharply the penalty acts, not the direction of the
trend.

For $\gamma$, the entropy variant's weakness is structural rather
than a tuning choice: the fine-tuned model's normalized entropy is
far below its maximum ($H/H_{\max} \approx \true{0.02}$ to
\true{0.07}, Table~\ref{tab:confidence}), so the entropy penalty
barely activates for any $\gamma \ge 1$.

\subsection{Computational Overhead}
\label{sec:appendix_overhead}

Constrained decoding adds negligible computational overhead ($<$5\% wall-clock time on average, Gemma 3 1B, single A100 GPU). The constraint logic (window lookup and logit subtraction) is trivially cheap; runtime variation is dominated by output length rather than the constraint computation itself.

\subsection{Why Static Wins Everywhere}
\label{sec:appendix_why_static}

On D1 (heterogeneous edits across a high-quality MT baseline), the pattern is consistent with an inconsistent training signal that mixes orthographic, stylistic, semantic, and structural corrections that vary unpredictably by sentence. The model nonetheless edits at high confidence whether or not the edit helps (Binary Collapse). Dynamic methods relax the constraint where the model is confident, but on D1 that confidence is systematically misplaced.

On D5 (genuine error correction), the explanation is subtler. The model \emph{does} learn useful edits (the $\lambda$ curve is U-shaped), yet the constraint ordering still holds: Static~$>$~PMT~$>$~Entropy. Here, static may win because useful corrections coexist with residual over-editing, and token-level confidence does not reliably distinguish the two. The edit-distance penalty does punish over-editing---every deviation from the MT window is penalized---but static applies it uniformly, whereas the dynamic variants waive it exactly where the model is confident, and confident tokens mix genuine corrections with confident over-edits in roughly equal measure. Static therefore suppresses residual over-edits regardless of a confidence signal that cannot separate the two. The D5 result indicates that miscalibration need not be specific to tasks that yield no net-positive edits. In our settings, it persists even when genuine corrections exist, alongside residual confidence-incoherent edits after fine-tuning on small heterogeneous corpora.

\subsection{Practical Guidelines}
\label{sec:appendix_practitioner}

The diagnostic is a single $\lambda$ sweep on a held-out validation
set with references. Run the static constraint over a range of
$\lambda$ (Table~\ref{tab:hyperparams}, extended upward until TER
flattens), record validation TER at each value, and read off two
things: the shape of the TER-vs-$\lambda$ curve and the location of
its minimum. The sweep costs one inference pass per $\lambda$ value
and adds negligible overhead (Appendix~\ref{sec:appendix_overhead}).
Shape and minimum location place a fine-tuned checkpoint into one of
four outcomes, each with a concrete next step
(Table~\ref{tab:sweep_outcomes}); the second signal, the constraint
ordering, then fixes which variant to deploy (Section~\ref{sec:action}).

\begin{table*}[!t]
\centering
\footnotesize
\setlength{\tabcolsep}{4pt}
\begin{tabularx}{\textwidth}{p{2.8cm}p{3.0cm}Xp{2.6cm}}
\toprule
\textbf{Curve signature} & \textbf{What it means} & \textbf{Recommended action} & \textbf{Example} \\
\midrule
U-shaped: minimum at $\lambda{>}0$, dipping below MT & Net-positive edits survived training & Deploy the static constraint at that $\lambda$ as a free inference-time gain (\S\ref{sec:action}) & D5; English--Marathi; Gemma~4B on D1 (marginal) \\
\addlinespace
Minimum at $\lambda{=}0$ & The model is already net-positive without any constraint & Deploy the vanilla fine-tuned checkpoint directly & D2 (News) \\
\addlinespace
Monotonic, reaching the MT baseline & No net-positive edits survived; copying is optimal & Copy MT for now; make the post-edits more consistent before training again & D1 (Literature) \\
\addlinespace
Monotonic, plateauing above the MT baseline & Over-editing is too severe to undo at inference & Address data or MT quality before training again & NLLB-600M on D1 \\
\bottomrule
\end{tabularx}
\caption{The four outcomes read off a single $\lambda$ sweep, with the next step each implies. The first two checkpoints are deployable as is; the last two call for upstream work on the training signal (\S\ref{sec:action}).}
\label{tab:sweep_outcomes}
\end{table*}

\section{Quality Estimation Details}
\label{sec:appendix_qe}

\paragraph{LaBSE semantic similarity.} Using LaBSE~\citep{feng-etal-2022-language}, we compute source--MT cosine similarity per test sample ($n{=}\true{716}$). Similarities are near-identical across edit categories (\true{0.881}/\true{0.877}/\true{0.865} for none/minor/major), the distributions overlapping almost completely; AUROCs for none vs.\ minor (\true{0.45}) and none vs.\ major (\true{0.42}) fall below chance. On D1, cross-lingual semantic similarity thus cannot distinguish sentences that need post-editing from those that do not.

\paragraph{COMET-QE.} We evaluate a state-of-the-art reference-free QE model, \texttt{Unbabel/\allowbreak wmt22-cometkiwi-da} \citep{rei-etal-2022-cometkiwi}, on D1 ($n{=}\true{716}$). Per-category scores are nearly flat (\true{0.868}/\true{0.870}/\true{0.843} for none/minor/major): AUROC~\true{0.58} (near chance), Spearman $r{=}\true{{-}0.18}$ with TER, and gating precision plateaus at \true{71\%} even at \true{95\%} recall, barely above the edit-rate prior (\true{71\%}). Despite training on millions of human quality judgments, COMET-QE cannot separate segments needing post-editing from those that do not, in this setting.

\paragraph{Fine-tuned QE formulations.} We fine-tune \texttt{xlm-roberta-large} \citep{conneau-etal-2020-unsupervised} on D1 ($n{=}\true{5{,}728}$) to predict TER from (source, MT) pairs, in three formulations: (i)~\emph{regression} on $\log(\text{TER}{+}1)$ (MSE loss), AUROC~\true{0.745}; (ii)~\emph{3-class} (none/minor/major) with inverse-frequency weights, AUROC~\true{0.703}, macro-F1~\true{0.43}; (iii)~\emph{binary} (edit-needed vs.\ not) with 50/50 oversampling, AUROC~\true{0.732}. The binary classifier gates at only \true{44\%} precision (default) and \true{53\%} at the 90th-percentile threshold, unsuitable for reliable gating. Multi-dataset training (adding D2, D3) degraded D1 performance. Table~\ref{tab:qe_results} summarizes all six approaches.

\begin{table}[!ht]
\centering
\small
\resizebox{\columnwidth}{!}{%
\begin{tabular}{lcc}
\toprule
\textbf{QE Approach} & \textbf{Best AUROC} & \textbf{Note} \\
\midrule
LaBSE (pretrained) & \true{0.45} & Below chance \\
COMET-QE (pretrained) & \true{0.58} & Near chance \\
XLM-R regression & \true{0.745} & Best overall \\
XLM-R 3-class & \true{0.703} & F1\textsubscript{macro}${=}\true{0.43}$ \\
XLM-R binary & \true{0.732} & Precision${=}\true{0.44}$ \\
\midrule
XLM-R (train on test) & \true{0.707} & Cannot memorize \\
\bottomrule
\end{tabular}}
\caption{QE approaches on D1. Fine-tuned models plateau at AUROC ${\sim}$\true{0.73}--\true{0.75}, insufficient for reliable gating.}
\label{tab:qe_results}
\end{table}

\section{Gradient Signal Analysis}
\label{sec:appendix_gradient}

To probe the mechanistic basis of Binary Collapse, we measure gradient signals from the fine-tuned Gemma~3 1B checkpoint (seed~42) on classified D1 and D5 samples, grouped by MQM category. Per sample we run one forward/backward pass (autocast off, LoRA parameters in float32), flatten the LoRA gradient into a single vector ($d{=}\true{26{,}091{,}520}$ across \true{364} LoRA tensors), and accumulate weighted group mean gradients, then take pairwise cosine similarities between them. Per-sample gradients are clipped at norm~\true{30.0}; those exceeding \true{1e6} are skipped (\true{0} on both datasets). The analysis uses one checkpoint and seed; replication across seeds and checkpoints is future work.

\subsection{D1 vs.\ D5: Gradient Norms}

Table~\ref{tab:gradient_norms} reports mean gradient norms and within-group loss variance by MQM category on D1 and D5. Three patterns hold on both. First, Style has the lowest mean gradient norm despite the largest sample count (D1: $n{=}\true{82}$, norm~\true{1.21}; D5: $n{=}\true{119}$, norm~\true{1.53}), consistent with within-group cancellation: its four subtypes (lexical preference, register shift, word order, loanword nativization) have meaningfully different per-sample norms (\true{8.1}--\true{10.3} on D1) and pull in different directions. Second, Style gives the highest within-group loss variance on D1 (coefficient of variation CV${=}\true{1.11}$ vs.\ \true{0.88} for Accuracy, \true{0.86} for Fluency) and is the only main group with CV${>}1.0$: per-sample surprise within Style exceeds its mean, the signature of an inconsistent signal. Third, Fluency shows the largest D1${\to}$D5 norm increase (\true{1.46}${\to}$\true{3.60}, $\Delta{=}\true{+2.14}$): on genuinely bad MT (D5) morphological and punctuation fixes are the clearest signal, whereas on D1 they are diluted by surrounding heterogeneous edits.

\begin{table}[!ht]
\centering
\small
\resizebox{\columnwidth}{!}{%
\begin{tabular}{lrcrrc}
\toprule
& \multicolumn{2}{c}{\textbf{D1 (heterogeneous)}}
& \phantom{x}
& \multicolumn{2}{c}{\textbf{D5 (genuine errors)}} \\
\cmidrule(lr){2-3} \cmidrule(lr){5-6}
\textbf{MQM Category}
& \textbf{$n$} & \textbf{Mean grad norm}
&
& \textbf{$n$} & \textbf{Mean grad norm} \\
\midrule
Accuracy      & \true{58}  & \true{1.58} && \true{32}  & \true{2.68} \\
Fluency       & \true{41}  & \true{1.46} && \true{19}  & \true{3.60} \\
Style         & \true{82}  & \true{1.21} && \true{119} & \true{1.53} \\
Retranslation & \true{17}  & \true{2.04} && \true{29}  & \true{2.97} \\
\bottomrule
\end{tabular}}
\caption{Mean gradient norms (norm of group mean LoRA gradient vector) by MQM category on D1 (heterogeneous edits) and D5 (genuine error correction). Loss CV (within-group coefficient of variation of per-sample loss) on D1: Accuracy \true{0.88}, Fluency \true{0.86}, Style \true{1.11}, Retranslation \true{1.64}.}
\label{tab:gradient_norms}
\end{table}

\subsection{D1 vs.\ D5: Cosine Similarity Matrices}

Table~\ref{tab:gradient_cosine} gives pairwise cosine similarities between group mean gradients on D1 and D5, and the contrast is sharp. On D1 the matrix is far from orthogonal (Style--Accuracy and maximum off-diagonal both \true{0.275}), consistent with Style and Accuracy gradients aligning under a shared copy prior despite different semantic content. On D5 every off-diagonal collapses toward zero (maximum \true{0.051}; Style--Accuracy \true{0.027}, a $10{\times}$ reduction) as each edit type pulls in its own direction. Fluency stays near-orthogonal on both (D1: \true{$\leq$0.019}; D5: \true{$\leq$0.029}): punctuation fixes are partially learned on D1 (Table~\ref{tab:edit_decomposition}), well-separated on D5. This mirrors the $\lambda$-curve contrast (Figure~\ref{fig:lambda_curves}): monotonic on D1, U-shaped on D5.

\begin{table}[!ht]
\centering
\small
\resizebox{\columnwidth}{!}{%
\begin{tabular}{lcccc|cccc}
\toprule
& \multicolumn{4}{c|}{\textbf{D1 (heterogeneous)}}
& \multicolumn{4}{c}{\textbf{D5 (genuine errors)}} \\
& \textbf{Acc.} & \textbf{Flu.} & \textbf{Sty.} & \textbf{Ret.}
& \textbf{Acc.} & \textbf{Flu.} & \textbf{Sty.} & \textbf{Ret.} \\
\midrule
\textbf{Accuracy}
  & 1.00 & \true{$-$0.01} & \true{0.28} & \true{0.14}
  & 1.00 & \true{$-$0.02} & \true{0.03} & \true{0.03} \\
\textbf{Fluency}
  & \true{$-$0.01} & 1.00 & \true{0.02} & \true{0.00}
  & \true{$-$0.02} & 1.00 & \true{0.03} & \true{0.00} \\
\textbf{Style}
  & \true{0.28} & \true{0.02} & 1.00 & \true{0.14}
  & \true{0.03} & \true{0.03} & 1.00 & \true{0.05} \\
\textbf{Retranslation}
  & \true{0.14} & \true{0.00} & \true{0.14} & 1.00
  & \true{0.03} & \true{0.00} & \true{0.05} & 1.00 \\
\bottomrule
\end{tabular}}
\caption{Pairwise cosine similarity between group mean LoRA gradients on D1 (left) and D5 (right), Gemma~3 1B, seed~42. On D1 the matrix is far from orthogonal; on D5 all off-diagonal entries approach zero (see text).}
\label{tab:gradient_cosine}
\end{table}

\section{Data-Level Mitigation: Edit-Type Filtering}
\label{sec:appendix_filter}

To test whether the failure is driven by edit-type composition rather than inconsistency, we retrained Gemma~3 1B on D1 under two matched-size conditions. \textbf{Filtered} removes samples GPT-5.2 classified as Style or Retranslation (Section~\ref{sec:edit_types}), leaving $n{=}\true{3{,}818}$ (${\sim}$67\% of D1's \true{5{,}728}); \textbf{Random control} is a size-matched random subsample (seed 42), isolating filtering from set size. Both reuse the main D1 hyperparameters (Table~\ref{tab:training_details}, seed 42) and the original test set ($n{=}\true{716}$).

\begin{table}[!ht]
\centering
\small
\begin{tabular}{lcc}
\toprule
\textbf{$\lambda$} & \textbf{Filtered} & \textbf{Random ctrl.} \\
\midrule
MT Base. & \true{24.98} & \true{24.98} \\
\midrule
0.0 & \true{26.74} & \true{26.37} \\
0.5 & \true{26.52} & \true{25.45} \\
1.0 & \true{26.46} & \true{25.15} \\
2.0 & \true{26.06} & \textbf{\true{24.97}} \\
3.0 & \true{26.06} & \textbf{\true{24.97}} \\
5.0 & \true{26.09} & \true{25.04} \\
7.0 & \true{26.08} & \true{25.04} \\
10.0 & \textbf{\true{25.00}} & \true{25.02} \\
\bottomrule
\end{tabular}
\caption{TER vs.\ $\lambda$ sweep on D1 (Gemma 3 1B, seed 42 only, static constraint; cross-seed replication pending) for the two matched-size retraining conditions of Appendix~\ref{sec:appendix_filter}: \textbf{Filtered} and \textbf{Random control}. $\lambda{=}0.0 \equiv$ vanilla fine-tuning; \textbf{MT Base.}\ is the shared MT baseline. Best (lowest) TER per condition in \textbf{bold}.}
\label{tab:filter_lambda}
\end{table}

Two findings emerge. First, the TER-vs-$\lambda$ curve (Table~\ref{tab:filter_lambda}) stays monotonically decreasing on filtered D1 (versus the U-shaped D5 curve in Figure~\ref{fig:lambda_curves}, center): no $\lambda$ crosses the MT baseline, so no net-positive edits survive. Second, filtering gives no advantage over the random control, whose TER stays about a point lower; where the control marginally reaches the baseline (best $\Delta$TER${=}\true{+0.01}$ at $\lambda{=}\true{2}$), the filtered set never does. Removing stylistic edits therefore does not improve outcomes: the residual Accuracy and Fluency edits have high within-group variance (Fluency CV~\true{0.86}, Accuracy CV~\true{0.88}), matching the cross-domain pattern (Table~\ref{tab:edit_cross_domain}) and the gradient analysis (Appendix~\ref{sec:appendix_gradient}). Filtering thus joins Appendix~\ref{sec:appendix_failed}'s failed interventions: none resets the diagnostic on D1.

\end{document}